\documentclass{article} % For LaTeX2e
\usepackage{colm2024_conference}
\usepackage{enumitem}

\usepackage{microtype}
\usepackage{hyperref}
\usepackage{url}
\usepackage{booktabs}
\usepackage{listings}
\usepackage{multirow}
\usepackage{graphicx}
\usepackage{subcaption}
\usepackage{multirow} % Multi rows in tables
\usepackage{makecell} % Multi cells in tables
\usepackage{arydshln} % Dashed lines in tables
\usepackage{bold-extra} % Bold face inside texttt environment

\usepackage{pifont}% http://ctan.org/pkg/pifont
\newcommand{\cmark}{\ding{51}}%
\newcommand{\xmark}{\ding{55}}%

\title{WorkBench: a Benchmark Dataset for Agents in a Realistic Workplace Setting}

\author{Olly Styles \thanks{Equal contributors}, Sam Miller \footnotemark[1], Patricio Cerda-Mardini  \\
Mindsdb
\And
Tanaya Guha \\
University of Glasgow
\AND
Victor Sanchez \\
University of Warwick
\AND
Bertie Vidgen \\
}

\colmfinalcopy % Uncomment for camera-ready version, but NOT for submission.
\begin{document}

\maketitle

\begin{abstract}
We introduce \textit{WorkBench}: a benchmark dataset for evaluating agents’ ability to execute tasks in a workplace setting. WorkBench contains a sandbox environment with five databases, 26 tools, and 690 tasks. These tasks represent common business activities, such as sending emails and scheduling meetings. The tasks in WorkBench are challenging as they require planning, tool selection, and often multiple actions. If a task has been successfully executed, one (or more) of the database values may change. The correct outcome for each task is unique and unambiguous, which allows for robust, automated evaluation. We call this key contribution \textit{outcome-centric evaluation}. We evaluate five existing ReAct agents on WorkBench, finding they successfully complete as few as 3\% of tasks (Llama2-70B), and just 43\% for the best-performing (GPT-4). We further find that agents’ errors can result in the wrong action being taken, such as an email being sent to the wrong person. WorkBench reveals weaknesses in agents’ ability to undertake common business activities, raising questions about their use in high-stakes workplace settings. WorkBench is publicly available as a free resource at \url{https://github.com/olly-styles/WorkBench}.\end{abstract}

\section{Introduction}
\label{intro}

\begin{figure}[b]
\begin{center}
   \includegraphics[width=\linewidth]{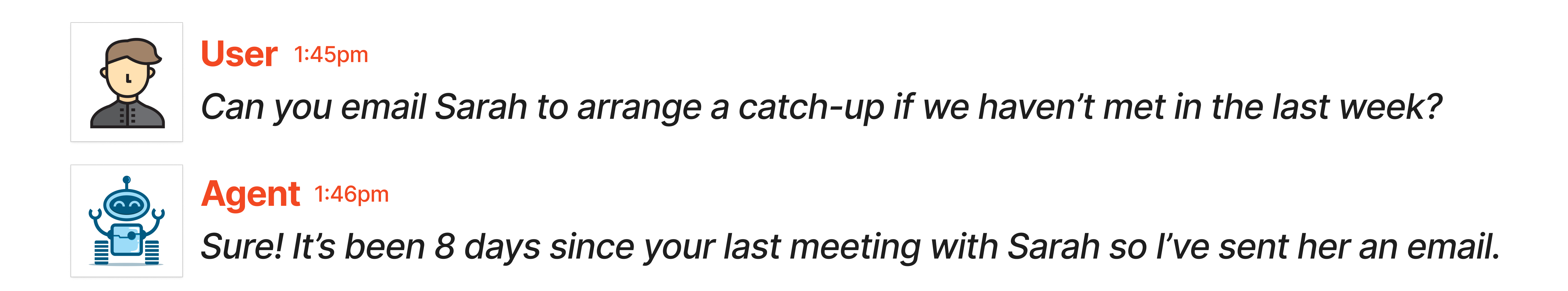}
\end{center}
\caption{\textbf{Agents in the workplace.} A sample task from WorkBench, the first dataset for evaluating autonomous agents on realistic workplace tasks.}
\label{fig:pull}
\end{figure}

Large language models (LLMs) excel at a broad range of tasks such as translation, summarisation and sentiment analysis. However, they often fail at tasks such as retrieving niche knowledge or recent information~\citep{bubeck2023sparks}. For example: they may give a wrong answer for this year's growth of the S\&P 500. Proposed approaches to overcoming these limitations include finetuning and Retrieval Augmented Generation~\citep{lewis2020}, or RAG, which gives LLMs access to data stored as embedded vectors. However, these approaches cannot deal with other LLM failures, such as simple arithmetic problems~\citep{brown2020few}. LLMs' inability to take actions also limits their capabilities. While they can summarise an email for a user, they cannot reply to it.

Autonomous agents, powered by LLMs with access to tools, can overcome these limitations. Tools are functions that agents use to interact with their environment. Rather than relying solely on knowledge from training, agents can use tools such as calculators and search engines to broaden their capabilities. With access to external systems, these agents can perform a range of tasks such as replying to emails and booking meetings.

While agents are promising, they make mistakes and little is known about their efficacy in practice. Recent benchmark datasets cover tasks like solving logic problems~\citep{mialon2023gaia}, or playing video games~\citep{wang2023voyager}. These help understand agents' broader capabilities, but do not represent tasks an agent could be asked to perform in a workplace setting.

This paper introduces \textit{WorkBench}: a benchmark dataset for evaluating agents in a realistic workplace setting. Tasks in WorkBench represent common functions across businesses, such as sending emails. WorkBench includes complex, multi-step tasks that require using several tools. For example: the task in Figure~\ref{fig:pull} requires an agent to review the user's calendar before sending an email. 

For tasks in WorkBench, we introduce \textit{outcome-centric evaluation}. Each task has a unique, unambiguous ground-truth outcome. Agents are evaluated on whether the outcome from their actions matches the ground truth outcome. This is a major step forward in enabling robust and automatic evaluation of action-taking agents.

We build a dataset of 690 unique tasks by combining human-curated task templates with a programmatic approach for creating multiple tasks per template. WorkBench tasks are challenging. We evaluate five different agents, finding that a state-of-the-art agent (ReAct with GPT-4) only completes 43\% of tasks correctly. 

To summarise, the contributions of this work are:
\begin{itemize}[leftmargin=*]
\itemsep0em 
  \item An opensource dataset, WorkBench, which enables robust and automatic evaluation of agents in a realistic workplace setting.
  \item Our outcome-centric evaluation methodology for benchmarking agents.
  \item Benchmark results showing what current state-of-the-art agents are capable of.
\end{itemize}

\section{Literature Review}
\label{lit_review}

\textbf{Tool Use Methods.}
\label{lit_review_methods}
Many prior works have investigated using LLMs as a planning agent for API-based tools (\cite{qin2023tool} provide a survey). Augmenting LLMs with tools enables them to interact with external systems, so end-users can interact with systems through a chat interface. Closed-source LLMs, like GPT-4~\citep{openai2023gpt4}, are assumed to be fine-tuned on tool-usage data, whereas opensource LLMs are generally not. This has led researchers to using closed-source LLMs to generate the training data for opensource LLMs~\citep{liu2023agentbench}.

\cite{yao2023react} propose the ReAct framework, which has been highly influential for agent design. This inspires approaches that i) transform the user’s intent into a high-level task, ii) make a plan to achieve the task based on available tools, and finally iii) execute that plan~\citep{qin2023toolllm, schick2023toolformer, bubeck2023sparks}. \cite{lu2023chameleon} show this approach is effective when using tools across multiple domains, such as maths and information retrieval, with GPT-4 as the LLM.

Many papers build on the insight that adding examples of tool use to the prompt may improve performance~\citep{schick2023toolformer}. \cite{hsieh2023tool} instead give tool documentation in the prompt, showing that this may be more effective than demonstrations. \cite{hao2023toolkengpt} overcome this by embedding each tool in a separate space, then using this embedding space to choose tools. \cite{song2023restgpt} show that first forming a natural language plan and then forming an explicit API call plan improves performance.

\textbf{Tool Use Limitations.}
%% Maybe get rid of this subsection if we need to save space.
A limitation of some studies is assessing tool usage with GPT-3.5 as the only LLM, so others have analysed the impact of choosing different LLMs~\citep{qin2023toolllm, hsieh2023tool, schick2023toolformer}.~\cite{ruan2023tptu} find that most smaller LLMs cannot effectively use tools, particularly for multi-step problems, with only GPT-3 effectively using tools. \cite{huang2023metatool} support this conclusion, finding that most LLMs other than GPT-4 cannot effectively select the right tools on their benchmark dataset. This may be explained by the finding of~\cite{xu2023tool} that GPT-4 can make correct API calls without in-context examples, whereas most models cannot. This suggests tool usage is part of the training data for GPT-4. Nevertheless, ~\cite{hao2023toolkengpt} propose improving opensource models by giving them access to embeddings of tool descriptions. \cite{ruan2024toolemu} study the risk of negative side effects when agents are used in real-world setting, which we build on in this work.

\textbf{Tool Use Evaluation.}
% There are many well-developed benchmark datasets for evaluating overall LLM performance, such as Google's Big-Bench~\citep{srivastava2023beyond}, Stanford's HELM~\citep{lee2023holistic} and EleutherAI's Harness~\citep{eval-harness}\footnote{This serves as the back-end for the popular HuggingFace OpenLLM Leaderboard}. These benchmarks feature hundreds of tasks for evaluating general LLM performance. However, they are of limited use for evaluating tool usage given that many of the tasks can be completed without access to tools.
There are many datasets for evaluating LLM capabilities, such as Big-Bench~\citep{srivastava2023beyond}, MMLU~\citep{mmlu} and HellaSwag~\citep{hellaswag}. These feature a range of topics such as coding, chess, and chemistry. However, they are of limited value for evaluating tool usage, given that questions are answerable without access to tools.

Other benchmarks have been proposed for specifically evaluating tool usage. \cite{mialon2023gaia} propose GAIA, which contains 466 questions that test agents on tasks such as web search and coding. Each question has a unique answer for robust, automatic evaluation. \cite{patil2023gorilla} present APIBench for evaluating the effectiveness of LLM agents in calling other AI models accessed through APIs. SLURP~\citep{bastianelli-etal-2020-slurp} and TaskMaster~\citep{byrne-etal-2019-taskmaster} focus on smart home tasks that only require a single action. Other benchmarks use simulated environments, such as playing video games~\citep{wang2023voyager}, human behaviour in a home~\citep{park2023generative}, and chat dialogues~\citep{budzianowski-etal-2018-multiwoz}.

\cite{xu2023tool} propose ToolBench for evaluating the efficacy of LLMs in constructing real API calls for tasks such as travel booking. They generate a large volume of tasks with ChatGPT. \cite{zhuang2023toolqa} propose ToolQA, which combines human task creation with a programmatic approach to scaling the dataset. However, both datasets are limited to information retrieval tasks. \cite{li2023apibank} propose API-Bank for evaluating agents using tools to take actions. Their automated approach enables a very large number of tools. However, answers are not unique or unambiguous. They require an LLM to evaluate performance of agents, therefore the validity of their evaluation depends on the evaluator LLM not making mistakes. Our proposed outcome-centric evaluation methodology overcomes this issue. Table~\ref{tab:dataset-comparision} compares WorkBench against existing datasets.

\textbf{Web Browsing Agents.}
Another relevant strand of literature is agents for web browsing. Their tools are implicitly defined via possible interactions with a browser \citep{mind2web2023, he2024webvoyagerbuildingendtoendweb, webshop2022}. Some web browsing benchmarks use text-based evaluation, which shares properties with our evaluation methodology\citep{zhou2024webarena, koh2024visualwebarenaevaluatingmultimodalagents}. Particularly relevant and concurrent with our work, \cite{drouin2024workarenacapablewebagents} also propose a robust method for evaluating retrieval tasks. Our benchmark builds on these papers by i) evaluating side effects from unintended state changes outside the target, and ii) focusing on workplace-relevant tasks.

\section{Proposed Benchmark Dataset: WorkBench}

\begin{table}[tb]
\setlength{\tabcolsep}{3pt} % Stretch table horizontally
\renewcommand*{\arraystretch}{1.25} % Stretch table vertical
  \begin{tabular}{l c c c c}
    \toprule
    \makecell{} & \bf Dataset & \bf Tool usage & \bf Actions & \bf \makecell{Outcome-centric \\ evaluation}\\ 
    \hline
    \multirow{3}{*}{\makecell{Datasets \\ without tools}} 
    & BigBench~\citep{srivastava2023beyond} & \xmark & \xmark & n/a\\
    & MMLU~\citep{mmlu} & \xmark & \xmark & n/a\\
    & HellaSwag~\citep{hellaswag} & \xmark & \xmark & n/a\\
    \noalign{\vskip 0.5mm}    \hdashline   \noalign{\vskip 0.5mm}    
    \multirow{3}{*}{\makecell{Retrieval-based\\ datasets}} 
    & GAIA~\citep{mialon2023gaia} & \cmark & \xmark & n/a\\
    & ToolQA~\citep{zhuang2023toolqa} & \cmark & \xmark & n/a\\ 
    & ToolBench~\citep{qin2023toolllm} & \cmark & \xmark & n/a\\ 
    \noalign{\vskip 0.5mm}    \hdashline    \noalign{\vskip 0.5mm}    
    \multirow{4}{*}{\makecell{Action-based\\ datasets}} & API-Bank~\citep{li2023apibank} & \cmark & \cmark & \xmark\\
    & WebArena~\citep{zhou2024webarena} & \cmark & \cmark & \xmark\\
    & WorkArena~\citep{drouin2024workarenacapablewebagents} & \cmark & \cmark & \xmark\\
    & WorkBench (Ours) & \cmark & \cmark & \cmark\\
    \bottomrule
    \vspace{-6.2mm}
  \end{tabular}
  \begin{center} \caption{\textbf{Dataset comparison.} WorkBench is the first dataset built with outcome-centric evaluation. This enables robust, automatic evaluation of agents on tasks requiring actions.}
  \label{tab:dataset-comparision}
  \end{center}
  \end{table}

\begin{figure}[h]
\begin{center}
   \includegraphics[width=\linewidth]{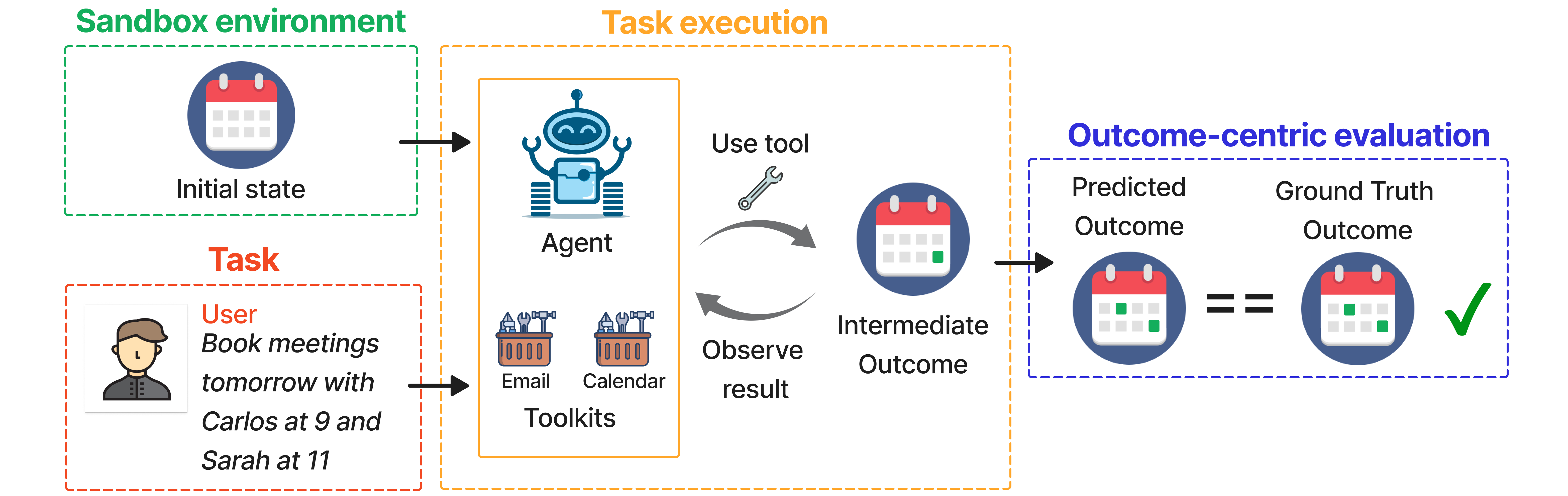}
\end{center}
   \caption{\textbf{Our complete pipeline for evaluating agents.} \textit{1)~Sandbox environment:} The sandbox has an initial state, defined by five databases. \textit{2)~Task}: a request is sent by the user. \textit{3)~Task execution:} a task is sent to the agent, which has access to toolkits in various domains. The agent takes actions using these tools, which may alter the sandbox databases. The agent observes the result of using the tool to determine if more actions are required. \textit{4)~Outcome-centric evaluation:} the updated sandbox databases are compared against the ground truth.}
\label{fig:pipeline}
\end{figure}

Figure~\ref{fig:pipeline} shows our complete methodology for evaluating agents. Real-world agents typically interact with data using external API calls. However, we do not use data from real APIs such as Gmail, as these require authentication and change over time. Instead, we create a local environment with five sandbox databases. These represent the initial state of the agent's environment. In response to a task, the agent's actions can alter the sandbox databases. All tasks in WorkBench have a unique, unambiguous outcome, which is the expected state of the sandbox databases after successful completion of the task.

\subsection{Sandbox Environment}
We create a sandbox environment consisting of five simulated databases:

\textbf{Calendar.} Each event has an ID, name, participant email, start time and duration. There are 300 events in the database. See Table~\ref{table:calendar_snippet} for a snippet of Calendar data. \newline
\textbf{Email.} Each email has an ID, sender, subject, body, and datetime. There are 500 emails in the database. \newline
\textbf{Website analytics.} Each website visit has a visitor ID, page view count, visit duration, traffic source, and visitor engagement score. There are 500 website visits in the database. \newline
\textbf{Customer relationship management (CRM).} Each customer has an ID, name, email, phone number, activity status, assigned employee, date of last contact, product interest, follow up deadline, and notes. There are 200 customers in our database. \newline
\textbf{Project management.} Each task has an ID, name, due date, assigned employee, list name, and board name. There are 300 tasks in our database.

\begin{table}[ht]
\setlength{\tabcolsep}{3pt} % Stretch table horizontally
\centering
\begin{tabular}{c c c c c}
\toprule
\textbf{Event ID}    & \textbf{Event Name}              & \textbf{Participant Email}   & \textbf{Start Time}       & \textbf{Duration Minutes} \\
\hline
000013 & \makecell{sync up}                             & luis.ortiz@atlas.com   & 2023-08-01 09:00 & 90 \\
000275 & \makecell{process review}                      & fatima.khan@atlas.com & 2023-08-01 11:30 & 30 \\
000264 & \makecell{Onboarding} & akira.sato@atlas.com  & 2023-08-01 14:30 & 60 \\
\bottomrule
\end{tabular}
\caption{\textbf{Calendar sandbox database snippet.} This is a sample of data from the Calendar sandbox database, which has a row for each event. We create sandbox databases for four other domains, which are further detailed in Appendix~\ref{appendix_simulated_data}}
\label{table:calendar_snippet}
\end{table}

\subsection{Task and outcome pairs}\label{sec:task_outcome_pairs}
We manually create task templates that represent realistic workplace tasks. Templates are split into two categories: 1) Single domain: these only require tools from one domain, such as Calendar, to complete. 2) Multi-domain: these require tools from multiple domains, such as Email and Calendar, to complete.

\begin{figure}[t]
\begin{center}
   \includegraphics[width=\linewidth]{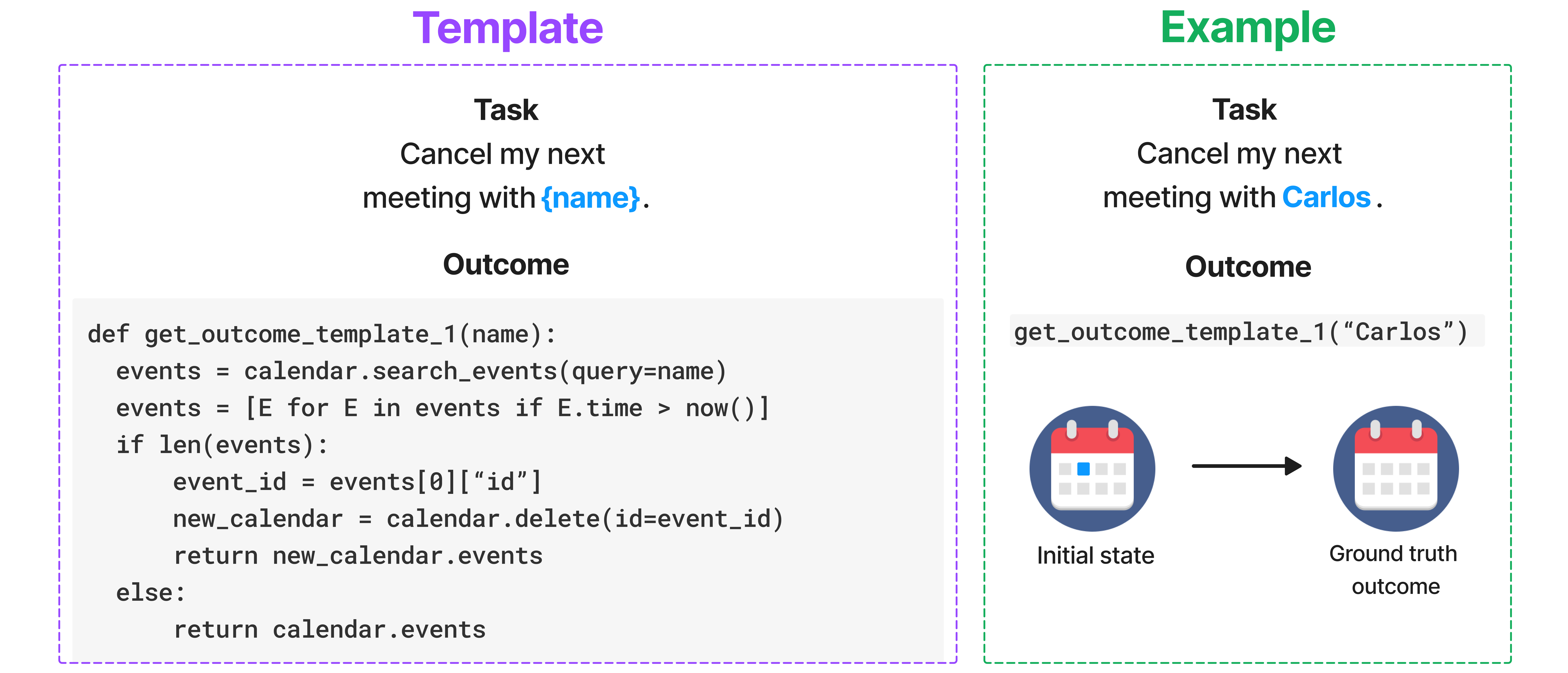}
\end{center}
   \caption{\textbf{Task and outcome creation.} The left side shows a pair of task-and-outcome templates. The outcome template is a function that returns the ground truth for the changes to the sandbox databases, given correct task completion. The right side shows a task-outcome pair created from these templates. In this example, the correct outcome is that the next meeting with Carlos is no longer in the Calendar sandbox.}
\label{fig:templates}
\end{figure}

Figure~\ref{fig:templates} shows how we programmatically create multiple tasks per template, and calculate the associated outcome. Across all domains, we create 10 unique tasks for each template, yielding 690 tasks in total. Table~\ref{table:unique_tasks_templates} shows the number of templates for each domain in WorkBench. Our templates contain added linguistic variation, so that each task is phrased in three different ways. We provide examples of this variation and empirical justification for our template approach in Appendix~\ref{appendix_template_dependence}.

\begin{table}[t]
\centering
\begin{tabular}{lcc}
\toprule
\textbf{Domain}     & \textbf{Number of unique templates} & \textbf{Number of unique tasks} \\
\noalign{\vskip 0.5mm}    \hline    \noalign{\vskip 0.5mm}    
Analytics & 12 & 120 \\
Calendar  & 11 & 110 \\
CRM  & 8 & 80 \\
Email     & 9 & 90 \\
Project Management   & 8 & 80 \\ 
Multi-domain     & 21 & 210 \\
\noalign{\vskip 0.5mm}    \hdashline    \noalign{\vskip 0.5mm}    
\textbf{Total} & 69 & 690 \\ 
\bottomrule
\end{tabular}
\caption{\textbf{Number of unique tasks and templates in each domain.} Our template approach creates 690 varied, realistic tasks. Each has an associated unique outcome.}
\label{table:unique_tasks_templates}
\end{table}

\begin{figure}[t]
\begin{center}
   \includegraphics[width=0.7\linewidth]{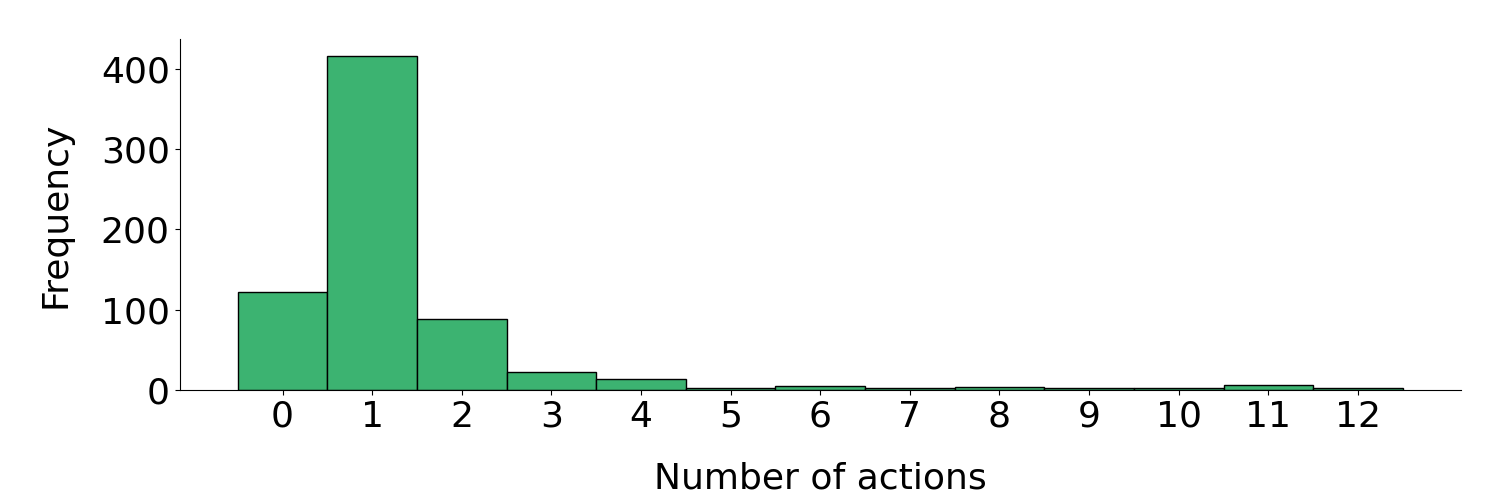}
\end{center}
   \caption{\textbf{Ground truth number of actions required to complete a task.} 18\% of tasks require no actions. Sometimes agent would use retrieval tools, but would not need to execute any actions. For example: a request to cancel meetings on a date when there aren't any scheduled. These tasks are easier, which we show in Appendix~\ref{appendix_no_action}.}
\label{fig:action_lengths}
\end{figure}

We create some very challenging task templates by combining actions across multiple domains. Figure~\ref{fig:action_lengths} shows the distribution of the number of actions required to complete each task in our database. Many tasks require multiple actions, with some requiring up to 12 actions. Here is an example of a complex multi-domain task:

\texttt{If our website page views fell by more than 10\% in the past week, schedule a 30-minute meeting with Sam called "Urgent Analytics Update" at the earliest free time tomorrow. Otherwise email them saying "Site traffic was stable the past week, nice work."}

We can define a simple function to automatically find the correct outcome. However, the agent must combine Analytics, Calendar and Email tools to complete this task. Appendix~\ref{appendix_example_tasks} provides more examples of tasks.

\subsection{Task Execution}

Agents execute tasks using 26 tools across the five domains, which are summarised in Table \ref{table:toolkits}. Each tool consists of a function that interacts with the sandbox databases, and documentation (a docstring) showing the agent how to use the tool. Each docstring contains a high-level description, parameters, return values, an example of tool usage, and any limits. An example limit is the \texttt{search\_events(start\_time, end\_time)}, which returns a maximum of five events. If the agent needs to find a large number of events, it must search multiple times with different time windows and then concatenate the results. Appendix \ref{appendix_prompt_and_tools} contains the full docstring for each tool.

\begin{table}[t]
\setlength{\tabcolsep}{3pt} % Stretch table horizontally
\centering
\small
\begin{tabular}{lllll}
\toprule
\textbf{Email} & \textbf{Calendar} & \textbf{Web Analytics} & \textbf{CRM} & \textbf{Projects} \\
\hline
\texttt{get\_email\_info} & \texttt{get\_event\_info} & \texttt{get\_visitor\_info} & \texttt{get\_customer\_info} & \texttt{get\_task\_info} \\

\texttt{search\_emails} & \texttt{search\_events} & \texttt{count\_traffic\_source} & \texttt{search\_customers} & \texttt{search\_tasks} \\

\texttt{send\_email} & \texttt{create\_event} & \texttt{count\_engaged\_users} & \texttt{update\_customer} & \texttt{create\_task} \\

\texttt{delete\_email} & \texttt{delete\_event} & \texttt{count\_total\_visits} & \texttt{add\_customer} & \texttt{delete\_task} \\

\texttt{forward\_email} & \texttt{update\_event} & \texttt{average\_visit\_duration} & \texttt{delete\_customer} & \texttt{update\_task} \\

\texttt{reply\_email} & & \texttt{create\_plot} & & \\
\bottomrule
\end{tabular}
   \caption{\textbf{Summary of toolkits.} We define 26 tools across 5 domains.}
\label{table:toolkits}
\end{table}

\subsection{Outcome-Centric Evaluation}

\begin{figure}[t]
\begin{center}
   \includegraphics[width=\linewidth]{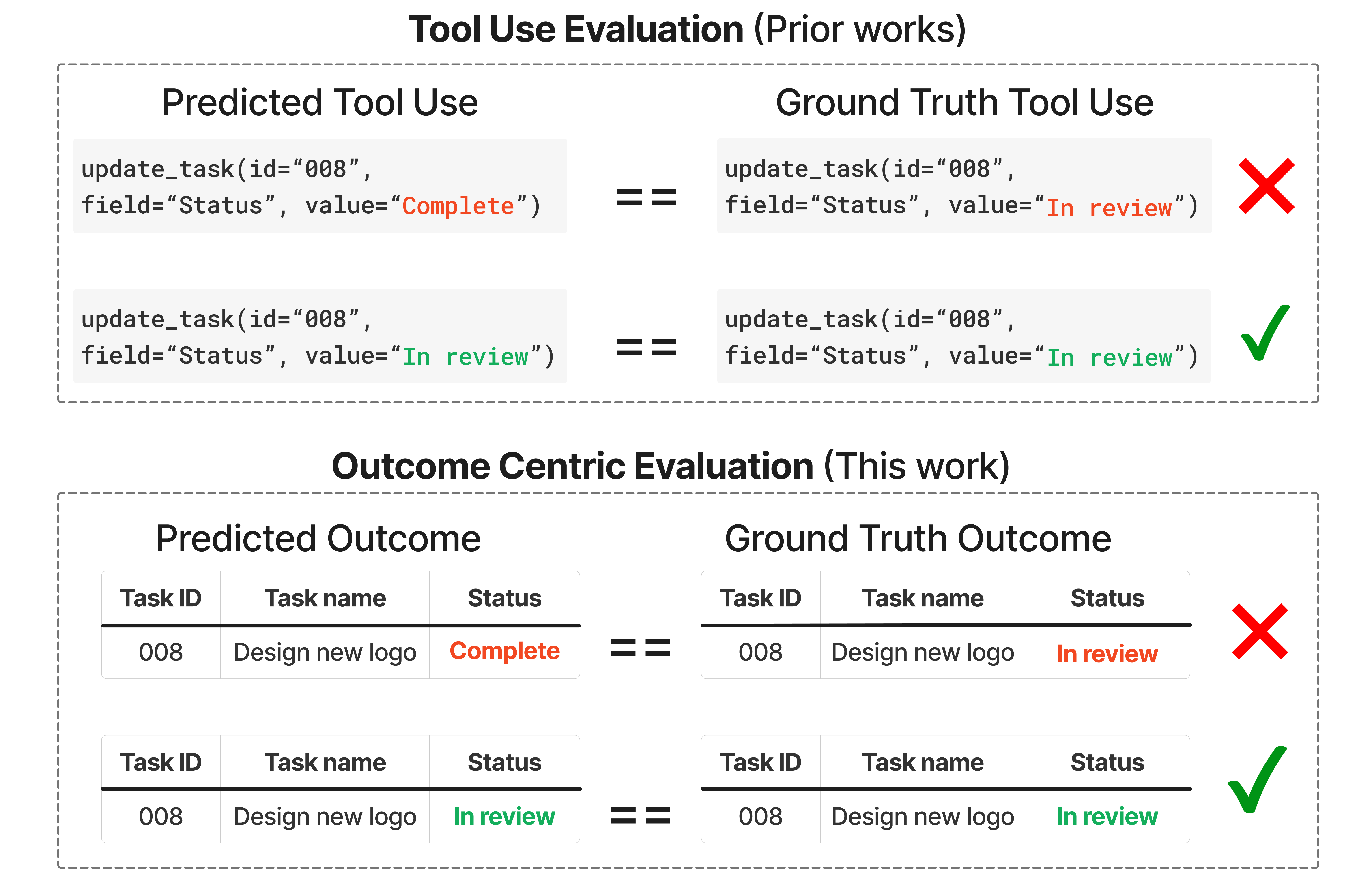}
\end{center}
   \caption{\textbf{Outcome-Centric Evaluation.} We propose outcome-centric evaluation, where there is a unique ground-truth outcome for each task (lower panel). We consider the task correctly executed if the predicted outcome following the agent's actions matches this outcome. This allows the agent to find multiple paths to the correct outcome, unlike prior works (upper panel) which evaluate the agent's function calls.}
\label{fig:eval}
\end{figure}

Figure~\ref{fig:templates} shows that the correct outcome, which is the ground truth, is always known. We then evaluate whether the outcome resulting from the agent's actions matches the ground truth. The ground truth therefore includes the state of all five sandbox databases. We call this methodology outcome-centric evaluation - Figure \ref{fig:eval} compares our outcome-centric evaluation against prior works, which evaluate the agent's function calls.

% We allow the agent to take any path of function calls if it leads to the correct outcome, rather than requiring the agent's function calls to match the expected function calls. For example:

% \texttt{Schedule a 30-minute meeting called "Analytics Update" on \textcolor{blue}{\{date\}} at the earliest time available with \textcolor{blue}{\{person\}}}

% The earliest time available is always known from the Calendar database. Therefore, this can be included as part of our outcome template, and the correct outcome will be automatically generated. However, the agent must find the earliest time available by combining Calendar tools with its internal reasoning. Once it has this time, it would need to schedule the event. A sensible path for the agent might be:

% \texttt{THOUGHT: I can find the earliest time available by searching events} \newline
% \texttt{ACTION: search\_events(date=2023-12-01)} \newline
% \texttt{OBSERVATION: \{9am - 10am meeting with Carlos, 2pm - 3pm meeting with Sam\}} \newline
% \texttt{THOUGHT: I should schedule the event at 10am, the first available slot.} \newline
% \texttt{ACTION: create\_event(participant="Kofi", date="2023-12-01", time="10:00", duration=30, event\_name="Analytics Update")} \newline
% \texttt{OBSERVATION: "Event created."} \newline
% \texttt{THOUGHT: The task has been completed.} \newline

An agent can follow any action path provided the resulting sandbox databases match the ground truth outcome. For example, sometimes the agent recovers from its error and takes the correct action: 

\textbf{Task:}\newline\textit{Make a task on the Front end board for Sam to improve conversion.}\newline
\textbf{Ground truth tool use:}\newline
\texttt{create\_task(name="improve conversion", board="Front end", assigned\_to="Sam")}\newline
\textbf{Agent's tool use:}\newline
\texttt{create\_task(name="improve conversion", board=\textcolor{red}{"Front End"}, assigned\_to="Sam")}\newline
\texttt{Observation: "`Front End' board does not exist, but `Front end' does..."}\newline
\texttt{create\_task(name="improve conversion", board="Front end", assigned\_to="Sam")}

Evaluation methods based on matching the function calls could unfairly find the agent had failed due to the extra calls. However, outcome-centric evaluation recognises that the agent was able to recover because the final change in state matches the ground truth outcome. As a result, the agent is not unfairly penalised.

Some prior benchmarks such as Gaia~\citep{mialon2023gaia} and ToolQA~\citep{zhuang2023toolqa} also have tasks with unique outcomes, but they are limited to information retrieval. WorkBench is the first dataset to evaluate tasks that require actions in this manner, due to our outcome-centric evaluation methodology.

\section{Results}

We assess the performance of LLM agents using the ReAct framework~\citep{yao2023react}. This enables the LLM to perform multiple action steps and update its action plan based on results from previous steps. 

\subsection{Performance Metrics}

Our primary metric is \textit{accuracy}. This is the \% of tasks where the outcome from the agent's actions match the expected outcome, which is the ground truth.

Our secondary metric is \textit{side effects}\footnote{This computer science term refers to a program altering variables outside its local environment.}. Some tools have negative consequences if used incorrectly, such as sending emails to the wrong person. If the agent's actions modify the sandbox databases in a way that does not match the ground truth exactly, we consider this a side effect. If the agent fails to complete the task, but does not alter the sandbox databases, then there are no side effects.

\subsection{Comparing Large Language Models}

Table~\ref{table:llm_comparison} compares five LLMs: GPT-3.5~\citep{brown2020few}, GPT-4~\citep{openai2023gpt4}, Claude-2~\citep{claude2}, Llama2-70B~\citep{llama2} and Mixtral-8x7B~\citep{mixtralofexperts}. GPT-4 greatly outperforms other models. For the worse-performing models, the main errors are insufficient context window length and failing to follow the ReAct framework.

\begin{table}[ht]
\setlength{\tabcolsep}{5pt} % Stretch table horizontally
\centering
\begin{tabular}{lccccc}
\toprule
 & \textbf{GPT-4} & \textbf{GPT-3.5} & \textbf{Claude-2} & \textbf{Llama2-70B} & \textbf{Mixtral-8x7B} \\
\noalign{\vskip 0.5mm}    \hline    \noalign{\vskip 0.5mm}    
Accuracy (required tools) & 49\% & 14\% & 23\% & 3\% & 20\% \\
Accuracy (all tools) & 43\% & 0\% & 26\% & 0\% & 16\% \\
\bottomrule
\end{tabular}
\caption{\textbf{Comparing accuracy across all models on WorkBench.} We selected these models based on general usage and prevalence in other benchmark studies. Due to limits on context window size, the upper row shows when models are given only the toolkit from the domain needed to complete each task. For example: if the task were "Book a 30-minute meeting with Sam for tomorrow at 9:30", we would provide all tools in the calendar toolkit and none from other toolkits. The lower row is when models are given all 26 tools. Scores of 0\% occur when the context window is not long enough to include all 26 tool descriptions.}
\label{table:llm_comparison}
\end{table}

Our benchmark is challenging for all models, including GPT-4. Given how poorly other models perform, we restrict further analysis to GPT-4. When giving this agent all 26 tools, rather than just the required toolkits, and find accuracy falls from 49\% to 43\%. This suggests the agent is negatively affected by redundant tools, which is consistent with prior studies~\citep{hao2023toolkengpt}. The next sections explore in further depth why the GPT-4 agent fails.

\subsection{Performance across domains}

Table~\ref{table:side_effects_across_domains} compares the GPT-4 agent's performance across our five individual domains, and tasks that require tools from multiple domains. Performance varies from 23\% accuracy on CRM tasks to 65\% for Calendar tasks. The agent is capable of combining tools across multiple domains. Its performance (40\%) on these tasks is similar to its average performance on single-domain tasks (43\%).

\begin{table}[h!]
\centering
\begin{tabular}{lcccccc}
\toprule
 & \multirow{2}{*}{\textbf{Analytics}} & \multirow{2}{*}{\textbf{Calendar}} & \multirow{2}{*}{\textbf{CRM}} & \multirow{2}{*}{\textbf{Email}} & \textbf{Project} & \textbf{Multi} \\
 &  &  &  &  & \textbf{Management} & \textbf{Domain} \\
\noalign{\vskip 0.5mm}    \hline    \noalign{\vskip 0.5mm}    
Number of tasks & 120 & 110 & 80 & 90 & 80 & 210 \\
\noalign{\vskip 0.5mm}    \hdashline    \noalign{\vskip 0.5mm}    
Accuracy (\textuparrow) & 39\% & 65\% & 23\% & 48\% & 39\% & 40\% \\
Side Effects (\textdownarrow) & 54\% & 22\% & 6\% & 6\% & 4\% & 29\% \\
\bottomrule
\end{tabular}
\caption{\textbf{Performance of GPT-4 ReAct agent on WorkBench with all tools provided.}. Higher is better (\textuparrow) for accuracy and lower is better (\textdownarrow) for side effects. Side effects are particularly common in the Analytics domain, as the agent often plots data for the wrong time period.}
\label{table:side_effects_across_domains}
\end{table}

% \begin{figure}[h]
% \begin{center}
%    \includegraphics[width=\linewidth]{./figs/number_of_tools_accuracy_comparison.png}
% \end{center}
%    \caption{\textbf{Impact of providing only the required toolkit(s) vs all toolkits for GPT-4 agent.} The number of tasks for each domain is shown in parenthesis. The agent's performance is robust up to providing redundant toolkits up to the maximum of 5 toolkits, consisting of 25 tools in total. However, its performance varies greatly across domains.}
% \label{fig:number_of_tools}
% \end{figure}

\subsection{Sources of Error}

\begin{figure}[t]
\centering
\includegraphics[width=\linewidth]{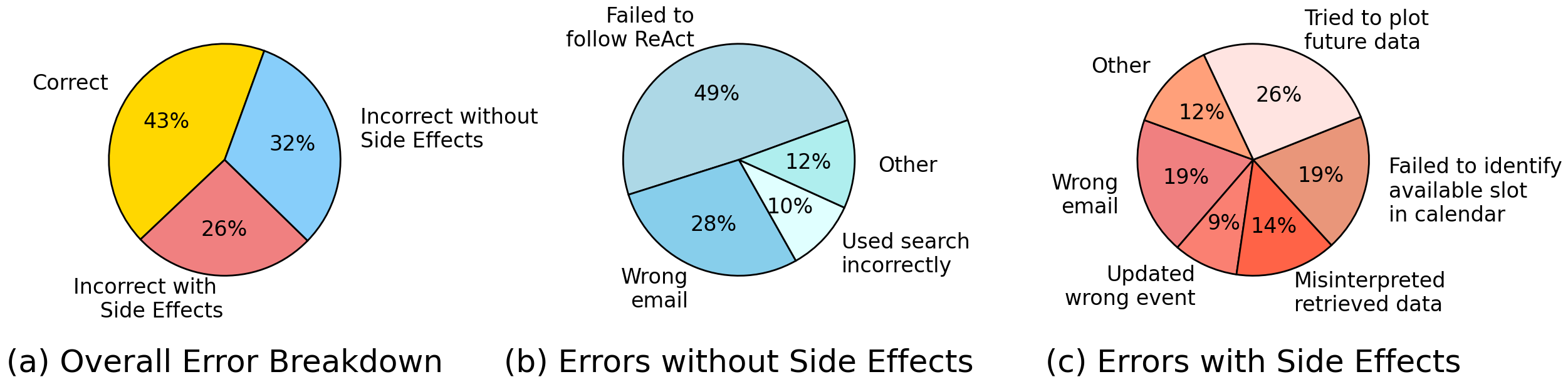}
\caption{\textbf{Error breakdown for GPT-4 across all domains with all tools provided.} Breakdown of errors with side effects vs those with no side effects (left). Detailed breakdown of errors with no side effects (middle) and errors with side effects (right).}
\label{fig:error_breakdown}
\end{figure}

Figure~\ref{fig:error_breakdown}a shows the prevalence of side effects. Side effects occur when the agent's actions modify the sandbox environment, but this change does not match the ground truth outcome. In this example from a task in our dataset, the agent cancels the wrong meeting:

\textbf{Task:} \textit{Cancel my next meeting with Nadia}\newline
\textbf{Ground truth:} \texttt{delete\_event(event\_id=00000035)}\newline
\textbf{Prediction:} \texttt{delete\_event(event\_id=\textcolor{red}{00000196})}

Errors without side effects occur when the agent fails to complete a task, but there are no unintended modifications to the environment. In this example, the agent does not take any actions because it searches on the wrong date:

\textit{(Today's date, Monday 20th November)}\newline
\textbf{Task}: Cancel all my meetings on Tuesday\newline
\textbf{Ground Truth}: \texttt{delete\_event(event\_id=00000025)}\newline
\textbf{Prediction}: Searches for meetings on 28th November - no meetings are found

Figures~\ref{fig:error_breakdown}b and \ref{fig:error_breakdown}c further break down errors by their most common sources. The most frequent error is failing to follow the ReAct framework. The agent must use the keyword \texttt{ACTION} followed by a \texttt{JSON} string with the tool name and arguments. The agent may omit the \texttt{ACTION} keyword, meaning no actions are performed. 

To address this source of error, we implement a re-sampling strategy. If the agent does not correctly generate the \texttt{ACTION} keyword, we repeat the same task up to 5 times. We use a temperature of 0.5 for retries, rather than the initial temperature of 0. With this strategy, the accuracy of the GPT-4 agent increases from 43\% to 49\%.

However, other common sources of error cannot be fixed with resampling. The agent often fails to find the correct email address when given names in the task. The agent may hallucinate an email address rather than using the search tool to find the correct email address. Here is an example, with intermediate steps hidden for concision:

\textbf{Task:} \textit{Forward all the emails from kofi last week about 'Staff Roster for Next Week' to fatima}\newline
\textbf{Ground truth:} \texttt{forward\_email(email\_id="0249",recipient="fatima.khan@atlas.com")} \newline
\textbf{Prediction:} \texttt{forward\_email(email\_id="0249", recipient="\textcolor{red}{fatima@example.com}")}

Errors also come from searching incorrectly. The agent may fail on tasks such as \textit{"Cancel my next meeting with Sam"} because it searches for events in the past. Similarly, the agent does not always account for the limits of its tools. It can fail on tasks such as \textit{"Cancel all future meetings with Sam"} because it only cancels a subset of future events after failing to factor in the limit on the number of results returned from searches. The agent could complete this task by repeating the search-and-deletion process, but fails to do so.

\section{Discussion and Future Work}

We have introduced WorkBench - the first benchmark that enables robust, automatic evaluation of agents in a workplace setting. Our approach ensures that each task has a unique outcome, which is the expected change to the state of the sandbox environment upon successful task completion.

One limitation of WorkBench is how well the sandbox environment represents real-world complexity. A real email inbox may contain tens of thousands of emails over many years, and emails are often spam, very long, and/or full of errors. Our initial results may therefore overestimate agents' current capabilities, so further studies could improve WorkBench by adding more challenging sandbox data.

While our tasks require multiple actions, they are limited to single-turn chat. Some longer queries may not represent realistic workplace interactions. Establishing a human baseline on our tasks would help determine the significance of this, but is hard to implement given our tasks would be completed by humans using a Graphical User Interface (GUI). Implementing a human baseline would measure GUI quality as well as task difficulty. Nevertheless, a multi-turn chat setup may be more representative of real tasks and could build upon our work.

We also found the agent's performance dropped when it had a greater number of tools it could choose from, but we were not able to explore this relationship fully with just 26 tools. Future work could extend our benchmark by adding more tools from other real-world domains such as HR software. This would help assess the relationship between agent accuracy and breadth of tooling.

A final limitation is that we do not assess pure retrieval tasks. While retrieval tools are a required intermediate step to complete many WorkBench tasks, we do not assess tasks that require solely retrieval such as finding a recent email. Future papers could build on our work with a method for assessing retrieval tasks, such as those proposed by other recent benchmarks \citep{drouin2024workarenacapablewebagents, zhuang2023toolqa}.

Despite these limitations, WorkBench has a large volume of high-quality, unique tasks. We include complex tasks that require planning, tool selection and analysing results across five domains. WorkBench is challenging, with the best agent achieving only 43\% accuracy. We find the main sources of error are the agent failing to execute its plan, giving the wrong arguments to tools, and not understanding the limits of its tools. Furthermore, errors often have negative consequences like sending emails to the wrong people. Future work could study fine-tuning LLMs to improve performance, such as~\citep{schick2023toolformer}.

WorkBench is both scalable and extensible. Future researchers could extend our dataset to include new domains, such as accounting tasks, and build an even larger dataset using our scalable approach to task creation. This will enable the evaluation of agents in progressively more complex settings as they continue to improve in the future.

% \subsubsection*{Author Contributions}
% Anonymous for now.

% \subsubsection*{Acknowledgments}
% Anonymous for now.

\bibliography{colm2024_conference}

\begin{thebibliography}{38}
\providecommand{\natexlab}[1]{#1}
\providecommand{\url}[1]{\texttt{#1}}
\expandafter\ifx\csname urlstyle\endcsname\relax
  \providecommand{\doi}[1]{doi: #1}\else
  \providecommand{\doi}{doi: \begingroup \urlstyle{rm}\Url}\fi

\bibitem[Anthropic(2023)]{claude2}
Anthropic.
\newblock Model card and evaluations for claude models.
\newblock 2023.
\newblock URL \url{https://www-cdn.anthropic.com/bd2a28d2535bfb0494cc8e2a3bf135d2e7523226/Model-Card-Claude-2.pdf}.

\bibitem[Bastianelli et~al.(2020)Bastianelli, Vanzo, Swietojanski, and Rieser]{bastianelli-etal-2020-slurp}
Emanuele Bastianelli, Andrea Vanzo, Pawel Swietojanski, and Verena Rieser.
\newblock {SLURP}: A spoken language understanding resource package.
\newblock In Bonnie Webber, Trevor Cohn, Yulan He, and Yang Liu (eds.), \emph{Proceedings of the 2020 Conference on Empirical Methods in Natural Language Processing (EMNLP)}, pp.\  7252--7262, Online, November 2020. Association for Computational Linguistics.
\newblock \doi{10.18653/v1/2020.emnlp-main.588}.
\newblock URL \url{https://aclanthology.org/2020.emnlp-main.588}.

\bibitem[Brown et~al.(2020)Brown, Mann, Ryder, Subbiah, Kaplan, Dhariwal, Neelakantan, Shyam, Sastry, Askell, Agarwal, Herbert-Voss, Krueger, Henighan, Child, Ramesh, Ziegler, Wu, Winter, Hesse, Chen, Sigler, Litwin, Gray, Chess, Clark, Berner, McCandlish, Radford, Sutskever, and Amodei]{brown2020few}
Tom Brown, Benjamin Mann, Nick Ryder, Melanie Subbiah, Jared~D Kaplan, Prafulla Dhariwal, Arvind Neelakantan, Pranav Shyam, Girish Sastry, Amanda Askell, Sandhini Agarwal, Ariel Herbert-Voss, Gretchen Krueger, Tom Henighan, Rewon Child, Aditya Ramesh, Daniel Ziegler, Jeffrey Wu, Clemens Winter, Chris Hesse, Mark Chen, Eric Sigler, Mateusz Litwin, Scott Gray, Benjamin Chess, Jack Clark, Christopher Berner, Sam McCandlish, Alec Radford, Ilya Sutskever, and Dario Amodei.
\newblock Language models are few-shot learners.
\newblock In H.~Larochelle, M.~Ranzato, R.~Hadsell, M.F. Balcan, and H.~Lin (eds.), \emph{Advances in Neural Information Processing Systems}, volume~33, pp.\  1877--1901. Curran Associates, Inc., 2020.
\newblock URL \url{https://proceedings.neurips.cc/paper_files/paper/2020/file/1457c0d6bfcb4967418bfb8ac142f64a-Paper.pdf}.

\bibitem[Bubeck et~al.(2023)Bubeck, Chandrasekaran, Eldan, Gehrke, Horvitz, Kamar, Lee, Lee, Li, Lundberg, Nori, Palangi, Ribeiro, and Zhang]{bubeck2023sparks}
Sébastien Bubeck, Varun Chandrasekaran, Ronen Eldan, Johannes Gehrke, Eric Horvitz, Ece Kamar, Peter Lee, Yin~Tat Lee, Yuanzhi Li, Scott Lundberg, Harsha Nori, Hamid Palangi, Marco~Tulio Ribeiro, and Yi~Zhang.
\newblock Sparks of artificial general intelligence: Early experiments with gpt-4, 2023.

\bibitem[Budzianowski et~al.(2018)Budzianowski, Wen, Tseng, Casanueva, Ultes, Ramadan, and Ga{\v{s}}i{\'c}]{budzianowski-etal-2018-multiwoz}
Pawe{\l} Budzianowski, Tsung-Hsien Wen, Bo-Hsiang Tseng, I{\~n}igo Casanueva, Stefan Ultes, Osman Ramadan, and Milica Ga{\v{s}}i{\'c}.
\newblock {M}ulti{WOZ} - a large-scale multi-domain {W}izard-of-{O}z dataset for task-oriented dialogue modelling.
\newblock In Ellen Riloff, David Chiang, Julia Hockenmaier, and Jun{'}ichi Tsujii (eds.), \emph{Proceedings of the 2018 Conference on Empirical Methods in Natural Language Processing}, pp.\  5016--5026, Brussels, Belgium, October-November 2018. Association for Computational Linguistics.
\newblock \doi{10.18653/v1/D18-1547}.
\newblock URL \url{https://aclanthology.org/D18-1547}.

\bibitem[Byrne et~al.(2019)Byrne, Krishnamoorthi, Sankar, Neelakantan, Goodrich, Duckworth, Yavuz, Dubey, Kim, and Cedilnik]{byrne-etal-2019-taskmaster}
Bill Byrne, Karthik Krishnamoorthi, Chinnadhurai Sankar, Arvind Neelakantan, Ben Goodrich, Daniel Duckworth, Semih Yavuz, Amit Dubey, Kyu-Young Kim, and Andy Cedilnik.
\newblock Taskmaster-1: Toward a realistic and diverse dialog dataset.
\newblock In Kentaro Inui, Jing Jiang, Vincent Ng, and Xiaojun Wan (eds.), \emph{Proceedings of the 2019 Conference on Empirical Methods in Natural Language Processing and the 9th International Joint Conference on Natural Language Processing (EMNLP-IJCNLP)}, pp.\  4516--4525, Hong Kong, China, November 2019. Association for Computational Linguistics.
\newblock \doi{10.18653/v1/D19-1459}.
\newblock URL \url{https://aclanthology.org/D19-1459}.

\bibitem[Deng et~al.(2023)Deng, Gu, Zheng, Chen, Stevens, Wang, Sun, and Su]{mind2web2023}
Xiang Deng, Yu~Gu, Boyuan Zheng, Shijie Chen, Sam Stevens, Boshi Wang, Huan Sun, and Yu~Su.
\newblock Mind2web: Towards a generalist agent for the web.
\newblock In A.~Oh, T.~Naumann, A.~Globerson, K.~Saenko, M.~Hardt, and S.~Levine (eds.), \emph{Advances in Neural Information Processing Systems}, volume~36, pp.\  28091--28114. Curran Associates, Inc., 2023.
\newblock URL \url{https://proceedings.neurips.cc/paper_files/paper/2023/file/5950bf290a1570ea401bf98882128160-Paper-Datasets_and_Benchmarks.pdf}.

\bibitem[Drouin et~al.(2024)Drouin, Gasse, Caccia, Laradji, Verme, Marty, Boisvert, Thakkar, Cappart, Vazquez, Chapados, and Lacoste]{drouin2024workarenacapablewebagents}
Alexandre Drouin, Maxime Gasse, Massimo Caccia, Issam~H. Laradji, Manuel~Del Verme, Tom Marty, Léo Boisvert, Megh Thakkar, Quentin Cappart, David Vazquez, Nicolas Chapados, and Alexandre Lacoste.
\newblock Workarena: How capable are web agents at solving common knowledge work tasks?
\newblock In \emph{The International Conference on Machine Learning}, 2024.

\bibitem[Hao et~al.(2023)Hao, Liu, Wang, and Hu]{hao2023toolkengpt}
Shibo Hao, Tianyang Liu, Zhen Wang, and Zhiting Hu.
\newblock Toolkengpt: Augmenting frozen language models with massive tools via tool embeddings, 2023.

\bibitem[He et~al.(2024)He, Yao, Ma, Yu, Dai, Zhang, Lan, and Yu]{he2024webvoyagerbuildingendtoendweb}
Hongliang He, Wenlin Yao, Kaixin Ma, Wenhao Yu, Yong Dai, Hongming Zhang, Zhenzhong Lan, and Dong Yu.
\newblock Webvoyager: Building an end-to-end web agent with large multimodal models, 2024.
\newblock URL \url{https://arxiv.org/abs/2401.13919}.

\bibitem[Hendrycks et~al.(2021)Hendrycks, Burns, Basart, Zou, Mazeika, Song, and Steinhardt]{mmlu}
Dan Hendrycks, Collin Burns, Steven Basart, Andy Zou, Mantas Mazeika, Dawn Song, and Jacob Steinhardt.
\newblock Measuring massive multitask language understanding.
\newblock \emph{Proceedings of the International Conference on Learning Representations (ICLR)}, 2021.

\bibitem[Hsieh et~al.(2023)Hsieh, Chen, Li, Fujii, Ratner, Lee, Krishna, and Pfister]{hsieh2023tool}
Cheng-Yu Hsieh, Si-An Chen, Chun-Liang Li, Yasuhisa Fujii, Alexander Ratner, Chen-Yu Lee, Ranjay Krishna, and Tomas Pfister.
\newblock Tool documentation enables zero-shot tool-usage with large language models, 2023.

\bibitem[Huang et~al.(2023)Huang, Shi, Li, Fan, Wu, Zhang, Liu, Zhou, Wan, Gong, and Sun]{huang2023metatool}
Yue Huang, Jiawen Shi, Yuan Li, Chenrui Fan, Siyuan Wu, Qihui Zhang, Yixin Liu, Pan Zhou, Yao Wan, Neil~Zhenqiang Gong, and Lichao Sun.
\newblock Metatool benchmark for large language models: Deciding whether to use tools and which to use, 2023.

\bibitem[Jiang et~al.(2024)Jiang, Sablayrolles, Roux, Mensch, Savary, Bamford, Chaplot, Casas, Hanna, Bressand, et~al.]{mixtralofexperts}
Albert~Q Jiang, Alexandre Sablayrolles, Antoine Roux, Arthur Mensch, Blanche Savary, Chris Bamford, Devendra~Singh Chaplot, Diego de~las Casas, Emma~Bou Hanna, Florian Bressand, et~al.
\newblock Mixtral of experts.
\newblock \emph{arXiv preprint arXiv:2401.04088}, 2024.

\bibitem[Koh et~al.(2024)Koh, Lo, Jang, Duvvur, Lim, Huang, Neubig, Zhou, Salakhutdinov, and Fried]{koh2024visualwebarenaevaluatingmultimodalagents}
Jing~Yu Koh, Robert Lo, Lawrence Jang, Vikram Duvvur, Ming~Chong Lim, Po-Yu Huang, Graham Neubig, Shuyan Zhou, Ruslan Salakhutdinov, and Daniel Fried.
\newblock Visualwebarena: Evaluating multimodal agents on realistic visual web tasks, 2024.
\newblock URL \url{https://arxiv.org/abs/2401.13649}.

\bibitem[Lewis et~al.(2020)Lewis, Perez, Piktus, Petroni, Karpukhin, Goyal, K\"{u}ttler, Lewis, Yih, Rockt\"{a}schel, Riedel, and Kiela]{lewis2020}
Patrick Lewis, Ethan Perez, Aleksandra Piktus, Fabio Petroni, Vladimir Karpukhin, Naman Goyal, Heinrich K\"{u}ttler, Mike Lewis, Wen-tau Yih, Tim Rockt\"{a}schel, Sebastian Riedel, and Douwe Kiela.
\newblock Retrieval-augmented generation for knowledge-intensive nlp tasks.
\newblock In H.~Larochelle, M.~Ranzato, R.~Hadsell, M.F. Balcan, and H.~Lin (eds.), \emph{Advances in Neural Information Processing Systems}, volume~33, pp.\  9459--9474. Curran Associates, Inc., 2020.
\newblock URL \url{https://proceedings.neurips.cc/paper_files/paper/2020/file/6b493230205f780e1bc26945df7481e5-Paper.pdf}.

\bibitem[Li et~al.(2023)Li, Zhao, Yu, Song, Li, Yu, Li, Huang, and Li]{li2023apibank}
Minghao Li, Yingxiu Zhao, Bowen Yu, Feifan Song, Hangyu Li, Haiyang Yu, Zhoujun Li, Fei Huang, and Yongbin Li.
\newblock Api-bank: A comprehensive benchmark for tool-augmented llms, 2023.

\bibitem[Liu et~al.(2023)Liu, Yu, Zhang, Xu, Lei, Lai, Gu, Ding, Men, Yang, Zhang, Deng, Zeng, Du, Zhang, Shen, Zhang, Su, Sun, Huang, Dong, and Tang]{liu2023agentbench}
Xiao Liu, Hao Yu, Hanchen Zhang, Yifan Xu, Xuanyu Lei, Hanyu Lai, Yu~Gu, Hangliang Ding, Kaiwen Men, Kejuan Yang, Shudan Zhang, Xiang Deng, Aohan Zeng, Zhengxiao Du, Chenhui Zhang, Sheng Shen, Tianjun Zhang, Yu~Su, Huan Sun, Minlie Huang, Yuxiao Dong, and Jie Tang.
\newblock Agentbench: Evaluating llms as agents, 2023.

\bibitem[Lu et~al.(2023)Lu, Peng, Cheng, Galley, Chang, Wu, Zhu, and Gao]{lu2023chameleon}
Pan Lu, Baolin Peng, Hao Cheng, Michel Galley, Kai-Wei Chang, Ying~Nian Wu, Song-Chun Zhu, and Jianfeng Gao.
\newblock Chameleon: Plug-and-play compositional reasoning with large language models.
\newblock In \emph{The 37th Conference on Neural Information Processing Systems (NeurIPS)}, 2023.

\bibitem[Mialon et~al.(2023)Mialon, Fourrier, Swift, Wolf, LeCun, and Scialom]{mialon2023gaia}
Grégoire Mialon, Clémentine Fourrier, Craig Swift, Thomas Wolf, Yann LeCun, and Thomas Scialom.
\newblock Gaia: a benchmark for general ai assistants, 2023.

\bibitem[OpenAI et~al.(2023)OpenAI, :, Achiam, Adler, Agarwal, Ahmad, Akkaya, Aleman, Almeida, Altenschmidt, Altman, Anadkat, Avila, Babuschkin, Balaji, Balcom, Baltescu, Bao, Bavarian, Belgum, Bello, Berdine, Bernadett-Shapiro, Berner, Bogdonoff, Boiko, Boyd, Brakman, Brockman, Brooks, Brundage, Button, Cai, Campbell, Cann, Carey, Carlson, Carmichael, Chan, Chang, Chantzis, Chen, Chen, Chen, Chen, Chen, Chess, Cho, Chu, Chung, Cummings, Currier, Dai, Decareaux, Degry, Deutsch, Deville, Dhar, Dohan, Dowling, Dunning, Ecoffet, Eleti, Eloundou, Farhi, Fedus, Felix, Fishman, Forte, Fulford, Gao, Georges, Gibson, Goel, Gogineni, Goh, Gontijo-Lopes, Gordon, Grafstein, Gray, Greene, Gross, Gu, Guo, Hallacy, Han, Harris, He, Heaton, Heidecke, Hesse, Hickey, Hickey, Hoeschele, Houghton, Hsu, Hu, Hu, Huizinga, Jain, Jain, Jang, Jiang, Jiang, Jin, Jin, Jomoto, Jonn, Jun, Kaftan, Łukasz Kaiser, Kamali, Kanitscheider, Keskar, Khan, Kilpatrick, Kim, Kim, Kim, Kirchner, Kiros, Knight, Kokotajlo, Łukasz Kondraciuk,
  Kondrich, Konstantinidis, Kosic, Krueger, Kuo, Lampe, Lan, Lee, Leike, Leung, Levy, Li, Lim, Lin, Lin, Litwin, Lopez, Lowe, Lue, Makanju, Malfacini, Manning, Markov, Markovski, Martin, Mayer, Mayne, McGrew, McKinney, McLeavey, McMillan, McNeil, Medina, Mehta, Menick, Metz, Mishchenko, Mishkin, Monaco, Morikawa, Mossing, Mu, Murati, Murk, Mély, Nair, Nakano, Nayak, Neelakantan, Ngo, Noh, Ouyang, O'Keefe, Pachocki, Paino, Palermo, Pantuliano, Parascandolo, Parish, Parparita, Passos, Pavlov, Peng, Perelman, de~Avila Belbute~Peres, Petrov, de~Oliveira~Pinto, Michael, Pokorny, Pokrass, Pong, Powell, Power, Power, Proehl, Puri, Radford, Rae, Ramesh, Raymond, Real, Rimbach, Ross, Rotsted, Roussez, Ryder, Saltarelli, Sanders, Santurkar, Sastry, Schmidt, Schnurr, Schulman, Selsam, Sheppard, Sherbakov, Shieh, Shoker, Shyam, Sidor, Sigler, Simens, Sitkin, Slama, Sohl, Sokolowsky, Song, Staudacher, Such, Summers, Sutskever, Tang, Tezak, Thompson, Tillet, Tootoonchian, Tseng, Tuggle, Turley, Tworek, Uribe, Vallone,
  Vijayvergiya, Voss, Wainwright, Wang, Wang, Wang, Ward, Wei, Weinmann, Welihinda, Welinder, Weng, Weng, Wiethoff, Willner, Winter, Wolrich, Wong, Workman, Wu, Wu, Wu, Xiao, Xu, Yoo, Yu, Yuan, Zaremba, Zellers, Zhang, Zhang, Zhao, Zheng, Zhuang, Zhuk, and Zoph]{openai2023gpt4}
OpenAI, :, Josh Achiam, Steven Adler, Sandhini Agarwal, Lama Ahmad, Ilge Akkaya, Florencia~Leoni Aleman, Diogo Almeida, Janko Altenschmidt, Sam Altman, Shyamal Anadkat, Red Avila, Igor Babuschkin, Suchir Balaji, Valerie Balcom, Paul Baltescu, Haiming Bao, Mo~Bavarian, Jeff Belgum, Irwan Bello, Jake Berdine, Gabriel Bernadett-Shapiro, Christopher Berner, Lenny Bogdonoff, Oleg Boiko, Madelaine Boyd, Anna-Luisa Brakman, Greg Brockman, Tim Brooks, Miles Brundage, Kevin Button, Trevor Cai, Rosie Campbell, Andrew Cann, Brittany Carey, Chelsea Carlson, Rory Carmichael, Brooke Chan, Che Chang, Fotis Chantzis, Derek Chen, Sully Chen, Ruby Chen, Jason Chen, Mark Chen, Ben Chess, Chester Cho, Casey Chu, Hyung~Won Chung, Dave Cummings, Jeremiah Currier, Yunxing Dai, Cory Decareaux, Thomas Degry, Noah Deutsch, Damien Deville, Arka Dhar, David Dohan, Steve Dowling, Sheila Dunning, Adrien Ecoffet, Atty Eleti, Tyna Eloundou, David Farhi, Liam Fedus, Niko Felix, Simón~Posada Fishman, Juston Forte, Isabella Fulford, Leo Gao,
  Elie Georges, Christian Gibson, Vik Goel, Tarun Gogineni, Gabriel Goh, Rapha Gontijo-Lopes, Jonathan Gordon, Morgan Grafstein, Scott Gray, Ryan Greene, Joshua Gross, Shixiang~Shane Gu, Yufei Guo, Chris Hallacy, Jesse Han, Jeff Harris, Yuchen He, Mike Heaton, Johannes Heidecke, Chris Hesse, Alan Hickey, Wade Hickey, Peter Hoeschele, Brandon Houghton, Kenny Hsu, Shengli Hu, Xin Hu, Joost Huizinga, Shantanu Jain, Shawn Jain, Joanne Jang, Angela Jiang, Roger Jiang, Haozhun Jin, Denny Jin, Shino Jomoto, Billie Jonn, Heewoo Jun, Tomer Kaftan, Łukasz Kaiser, Ali Kamali, Ingmar Kanitscheider, Nitish~Shirish Keskar, Tabarak Khan, Logan Kilpatrick, Jong~Wook Kim, Christina Kim, Yongjik Kim, Hendrik Kirchner, Jamie Kiros, Matt Knight, Daniel Kokotajlo, Łukasz Kondraciuk, Andrew Kondrich, Aris Konstantinidis, Kyle Kosic, Gretchen Krueger, Vishal Kuo, Michael Lampe, Ikai Lan, Teddy Lee, Jan Leike, Jade Leung, Daniel Levy, Chak~Ming Li, Rachel Lim, Molly Lin, Stephanie Lin, Mateusz Litwin, Theresa Lopez, Ryan Lowe,
  Patricia Lue, Anna Makanju, Kim Malfacini, Sam Manning, Todor Markov, Yaniv Markovski, Bianca Martin, Katie Mayer, Andrew Mayne, Bob McGrew, Scott~Mayer McKinney, Christine McLeavey, Paul McMillan, Jake McNeil, David Medina, Aalok Mehta, Jacob Menick, Luke Metz, Andrey Mishchenko, Pamela Mishkin, Vinnie Monaco, Evan Morikawa, Daniel Mossing, Tong Mu, Mira Murati, Oleg Murk, David Mély, Ashvin Nair, Reiichiro Nakano, Rajeev Nayak, Arvind Neelakantan, Richard Ngo, Hyeonwoo Noh, Long Ouyang, Cullen O'Keefe, Jakub Pachocki, Alex Paino, Joe Palermo, Ashley Pantuliano, Giambattista Parascandolo, Joel Parish, Emy Parparita, Alex Passos, Mikhail Pavlov, Andrew Peng, Adam Perelman, Filipe de~Avila Belbute~Peres, Michael Petrov, Henrique~Ponde de~Oliveira~Pinto, Michael, Pokorny, Michelle Pokrass, Vitchyr Pong, Tolly Powell, Alethea Power, Boris Power, Elizabeth Proehl, Raul Puri, Alec Radford, Jack Rae, Aditya Ramesh, Cameron Raymond, Francis Real, Kendra Rimbach, Carl Ross, Bob Rotsted, Henri Roussez, Nick Ryder,
  Mario Saltarelli, Ted Sanders, Shibani Santurkar, Girish Sastry, Heather Schmidt, David Schnurr, John Schulman, Daniel Selsam, Kyla Sheppard, Toki Sherbakov, Jessica Shieh, Sarah Shoker, Pranav Shyam, Szymon Sidor, Eric Sigler, Maddie Simens, Jordan Sitkin, Katarina Slama, Ian Sohl, Benjamin Sokolowsky, Yang Song, Natalie Staudacher, Felipe~Petroski Such, Natalie Summers, Ilya Sutskever, Jie Tang, Nikolas Tezak, Madeleine Thompson, Phil Tillet, Amin Tootoonchian, Elizabeth Tseng, Preston Tuggle, Nick Turley, Jerry Tworek, Juan Felipe~Cerón Uribe, Andrea Vallone, Arun Vijayvergiya, Chelsea Voss, Carroll Wainwright, Justin~Jay Wang, Alvin Wang, Ben Wang, Jonathan Ward, Jason Wei, CJ~Weinmann, Akila Welihinda, Peter Welinder, Jiayi Weng, Lilian Weng, Matt Wiethoff, Dave Willner, Clemens Winter, Samuel Wolrich, Hannah Wong, Lauren Workman, Sherwin Wu, Jeff Wu, Michael Wu, Kai Xiao, Tao Xu, Sarah Yoo, Kevin Yu, Qiming Yuan, Wojciech Zaremba, Rowan Zellers, Chong Zhang, Marvin Zhang, Shengjia Zhao, Tianhao
  Zheng, Juntang Zhuang, William Zhuk, and Barret Zoph.
\newblock Gpt-4 technical report, 2023.

\bibitem[Park et~al.(2023)Park, O'Brien, Cai, Morris, Liang, and Bernstein]{park2023generative}
Joon~Sung Park, Joseph~C. O'Brien, Carrie~J. Cai, Meredith~Ringel Morris, Percy Liang, and Michael~S. Bernstein.
\newblock Generative agents: Interactive simulacra of human behavior, 2023.

\bibitem[Patil et~al.(2023)Patil, Zhang, Wang, and Gonzalez]{patil2023gorilla}
Shishir~G. Patil, Tianjun Zhang, Xin Wang, and Joseph~E. Gonzalez.
\newblock Gorilla: Large language model connected with massive apis, 2023.

\bibitem[Qin et~al.(2023{\natexlab{a}})Qin, Hu, Lin, Chen, Ding, Cui, Zeng, Huang, Xiao, Han, Fung, Su, Wang, Qian, Tian, Zhu, Liang, Shen, Xu, Zhang, Ye, Li, Tang, Yi, Zhu, Dai, Yan, Cong, Lu, Zhao, Huang, Yan, Han, Sun, Li, Phang, Yang, Wu, Ji, Liu, and Sun]{qin2023tool}
Yujia Qin, Shengding Hu, Yankai Lin, Weize Chen, Ning Ding, Ganqu Cui, Zheni Zeng, Yufei Huang, Chaojun Xiao, Chi Han, Yi~Ren Fung, Yusheng Su, Huadong Wang, Cheng Qian, Runchu Tian, Kunlun Zhu, Shihao Liang, Xingyu Shen, Bokai Xu, Zhen Zhang, Yining Ye, Bowen Li, Ziwei Tang, Jing Yi, Yuzhang Zhu, Zhenning Dai, Lan Yan, Xin Cong, Yaxi Lu, Weilin Zhao, Yuxiang Huang, Junxi Yan, Xu~Han, Xian Sun, Dahai Li, Jason Phang, Cheng Yang, Tongshuang Wu, Heng Ji, Zhiyuan Liu, and Maosong Sun.
\newblock Tool learning with foundation models, 2023{\natexlab{a}}.

\bibitem[Qin et~al.(2023{\natexlab{b}})Qin, Liang, Ye, Zhu, Yan, Lu, Lin, Cong, Tang, Qian, Zhao, Hong, Tian, Xie, Zhou, Gerstein, Li, Liu, and Sun]{qin2023toolllm}
Yujia Qin, Shihao Liang, Yining Ye, Kunlun Zhu, Lan Yan, Yaxi Lu, Yankai Lin, Xin Cong, Xiangru Tang, Bill Qian, Sihan Zhao, Lauren Hong, Runchu Tian, Ruobing Xie, Jie Zhou, Mark Gerstein, Dahai Li, Zhiyuan Liu, and Maosong Sun.
\newblock Toolllm: Facilitating large language models to master 16000+ real-world apis, 2023{\natexlab{b}}.

\bibitem[Ruan et~al.(2023)Ruan, Chen, Zhang, Xu, Bao, Du, Shi, Mao, Li, Zeng, and Zhao]{ruan2023tptu}
Jingqing Ruan, Yihong Chen, Bin Zhang, Zhiwei Xu, Tianpeng Bao, Guoqing Du, Shiwei Shi, Hangyu Mao, Ziyue Li, Xingyu Zeng, and Rui Zhao.
\newblock Tptu: Large language model-based ai agents for task planning and tool usage, 2023.

\bibitem[Ruan et~al.(2024)Ruan, Dong, Wang, Pitis, Zhou, Ba, Dubois, Maddison, and Hashimoto]{ruan2024toolemu}
Yangjun Ruan, Honghua Dong, Andrew Wang, Silviu Pitis, Yongchao Zhou, Jimmy Ba, Yann Dubois, Chris~J Maddison, and Tatsunori Hashimoto.
\newblock Identifying the risks of lm agents with an lm-emulated sandbox.
\newblock In \emph{The Twelfth International Conference on Learning Representations}, 2024.

\bibitem[Schick et~al.(2023)Schick, Dwivedi-Yu, Dessì, Raileanu, Lomeli, Zettlemoyer, Cancedda, and Scialom]{schick2023toolformer}
Timo Schick, Jane Dwivedi-Yu, Roberto Dessì, Roberta Raileanu, Maria Lomeli, Luke Zettlemoyer, Nicola Cancedda, and Thomas Scialom.
\newblock Toolformer: Language models can teach themselves to use tools, 2023.

\bibitem[Song et~al.(2023)Song, Xiong, Zhu, Wu, Qian, Song, Huang, Li, Wang, Yao, Tian, and Li]{song2023restgpt}
Yifan Song, Weimin Xiong, Dawei Zhu, Wenhao Wu, Han Qian, Mingbo Song, Hailiang Huang, Cheng Li, Ke~Wang, Rong Yao, Ye~Tian, and Sujian Li.
\newblock Restgpt: Connecting large language models with real-world restful apis, 2023.

\bibitem[{Srivavastava et. al}(2023)]{srivastava2023beyond}
{Srivavastava et. al}.
\newblock Beyond the imitation game: Quantifying and extrapolating the capabilities of language models.
\newblock \emph{Transactions on Machine Learning Research}, 2023.
\newblock ISSN 2835-8856.
\newblock URL \url{https://openreview.net/forum?id=uyTL5Bvosj}.

\bibitem[Touvron et~al.(2023)Touvron, Martin, Stone, Albert, Almahairi, Babaei, Bashlykov, Batra, Bhargava, Bhosale, et~al.]{llama2}
Hugo Touvron, Louis Martin, Kevin Stone, Peter Albert, Amjad Almahairi, Yasmine Babaei, Nikolay Bashlykov, Soumya Batra, Prajjwal Bhargava, Shruti Bhosale, et~al.
\newblock Llama 2: Open foundation and fine-tuned chat models.
\newblock \emph{arXiv preprint arXiv:2307.09288}, 2023.

\bibitem[Wang et~al.(2023)Wang, Xie, Jiang, Mandlekar, Xiao, Zhu, Fan, and Anandkumar]{wang2023voyager}
Guanzhi Wang, Yuqi Xie, Yunfan Jiang, Ajay Mandlekar, Chaowei Xiao, Yuke Zhu, Linxi Fan, and Anima Anandkumar.
\newblock Voyager: An open-ended embodied agent with large language models, 2023.

\bibitem[Xu et~al.(2023)Xu, Hong, Li, Hu, Chen, and Zhang]{xu2023tool}
Qiantong Xu, Fenglu Hong, Bo~Li, Changran Hu, Zhengyu Chen, and Jian Zhang.
\newblock On the tool manipulation capability of open-source large language models, 2023.

\bibitem[Yao et~al.(2022)Yao, Chen, Yang, and Narasimhan]{webshop2022}
Shunyu Yao, Howard Chen, John Yang, and Karthik Narasimhan.
\newblock Webshop: Towards scalable real-world web interaction with grounded language agents.
\newblock In S.~Koyejo, S.~Mohamed, A.~Agarwal, D.~Belgrave, K.~Cho, and A.~Oh (eds.), \emph{Advances in Neural Information Processing Systems}, volume~35, pp.\  20744--20757. Curran Associates, Inc., 2022.
\newblock URL \url{https://proceedings.neurips.cc/paper_files/paper/2022/file/82ad13ec01f9fe44c01cb91814fd7b8c-Paper-Conference.pdf}.

\bibitem[Yao et~al.(2023)Yao, Zhao, Yu, Du, Shafran, Narasimhan, and Cao]{yao2023react}
Shunyu Yao, Jeffrey Zhao, Dian Yu, Nan Du, Izhak Shafran, Karthik Narasimhan, and Yuan Cao.
\newblock React: Synergizing reasoning and acting in language models, 2023.

\bibitem[Zellers et~al.(2019)Zellers, Holtzman, Bisk, Farhadi, and Choi]{hellaswag}
Rowan Zellers, Ari Holtzman, Yonatan Bisk, Ali Farhadi, and Yejin Choi.
\newblock Hellaswag: Can a machine really finish your sentence?
\newblock In \emph{Proceedings of the 57th Annual Meeting of the Association for Computational Linguistics}. Association for Computational Linguistics, 2019.

\bibitem[Zhou et~al.(2024)Zhou, Xu, Zhu, Zhou, Lo, Sridhar, Cheng, Ou, Bisk, Fried, Alon, and Neubig]{zhou2024webarena}
Shuyan Zhou, Frank~F. Xu, Hao Zhu, Xuhui Zhou, Robert Lo, Abishek Sridhar, Xianyi Cheng, Tianyue Ou, Yonatan Bisk, Daniel Fried, Uri Alon, and Graham Neubig.
\newblock Webarena: A realistic web environment for building autonomous agents.
\newblock In \emph{The Twelfth International Conference on Learning Representations}, 2024.
\newblock URL \url{https://openreview.net/forum?id=oKn9c6ytLx}.

\bibitem[Zhuang et~al.(2023)Zhuang, Yu, Wang, Sun, and Zhang]{zhuang2023toolqa}
Yuchen Zhuang, Yue Yu, Kuan Wang, Haotian Sun, and Chao Zhang.
\newblock Toolqa: A dataset for llm question answering with external tools, 2023.

\end{thebibliography}
\bibliographystyle{colm2024_conference}

\appendix
\section{Appendix}
\subsection{Simulated database creation}\label{appendix_simulated_data}
Below is a sample of 5 rows from each of our 5 sandbox databases.

\subsubsection{Analytics}
\texttt{\textbf{date\_of\_visit,visitor\_id,page\_views,session\_duration\_seconds,traffic\_source,user\_engaged}\newline
2023-10-22,860,8,4,referral,False\newline
2023-11-22, 214, 11, 1, search engine, False\newline
2023-09-24, 130, 18, 0, social media, False\newline
2023-10-08, 385, 2, 4, direct, False\newline
2023-09-22, 252, 2, 11, search engine, False}

\subsubsection{Calendar}
\texttt{event\_id, event\_name, participant\_email, event\_start, duration\newline
00000013, sync up, luis.ortiz@atlas.com, 2023-08-01 09:00:00, 90\newline
00000275, process review, fatima.khan@atlas.com, 2023-08-01 11:30:00, 90\newline
00000098, Data Security and Compliance Training, amir.ali@atlas.com, 2023-08-02 11:00:00, 30\newline
00000190, Product Launch Analysis, yuki.tanaka@atlas.com, 2023-08-02 11:30:00, 30\newline
00000071, daily stand-up, kofi.mensah@atlas.com, 2023-08-02 13:30:00, 30\newline
}

\subsubsection{Customer Relationship Manager}
\texttt{\textbf{customer\_id, assigned\_to\_email, customer\_name, customer\_email, customer\_phone, last\_contact\_date, product\_interest, status, follow\_up\_by, notes}\newline
00000189, lena.schmidt@atlas.com, Taylor Jackson, taylor.jackson@nanolabs, , 2023-11-30, Consulting, Lost, 2023-12-22, 2023-11-07: Had a call. 2023-11-25: Had a call. \newline
00000107, sofia.santos@atlas.com, Quinn Harris, quinn.harris@nanoforcerobotics, , 2023-11-30, Consulting, Proposal, 2023-12-14, 2023-11-26: Saw the demo. 2023-11-29: Had a call. 2023-10-27: Had a call. \newline
00000052, raj.patel@atlas.com, Jaden White, jaden.white@protracefoods, 724-857-2625, 2023-11-30, Hardware, Won, 2023-12-13, 2023-10-17: Had a call. \newline
00000102, sofia.santos@atlas.com, Alex Thomas, alex.thomas@proenergy, , 2023-11-30, Hardware, Qualified, 2023-12-22, 2023-10-15: On holiday. \newline
00000187, lena.schmidt@atlas.com, Quinn Robinson, quinn.robinson@flexenergy, 399-396-5380, 2023-11-30, Hardware, Lead, 2023-12-23, }

\subsubsection{Email}
\texttt{\textbf{email\_id, inbox/outbox, sender/recipient, subject, sent\_datetime, body} \newline
00000373, inbox, santiago.martinez@atlas.com, Task Update on Develop prototype for payment gateway, 2023-10-01 09:15:02, "Sam, \textbackslash n Completed task 'Develop prototype for payment gateway' ahead of schedule. Please review and let me know if any tweaks are needed.\textbackslash n\textbackslash n Best,\textbackslash n Santiago"\newline
00000353, inbox, chenwei.zhang@atlas.com, Update on Annual Budget Planning Session, 2023-10-01 09:40:01, "Sam,\textbackslash n Encountered a few challenges while working on the Annual Budget Planning Session. Could use your advice.\textbackslash n\textbackslash n Cheers,\textbackslash n Chenwei"
00000013, inbox, kofi.mensah@atlas.com, Task Update on Fix alignment issue in homepage, 2023-10-01 10:50:46, "Dear Sam, \textbackslash n Regarding task 'Fix alignment issue in homepage',  I've made significant progress but have hit a snag with third-party API compatibility. Could use a brainstorm session.\textbackslash n \textbackslash n Regards, \textbackslash n Kofi"\newline
00000103, inbox, chenwei.zhang@atlas.com, Update on Quarterly Sales Review, 2023-10-01 10:58:07, "Hey Sam, \textbackslash n Encountered a few challenges while working on the Quarterly Sales Review. Could use your advice.\textbackslash n \textbackslash n Thanks, \textbackslash n Chenwei"\newline
00000295, inbox, nadia.moreau@atlas.com, Update on Year-End Performance Assessment, 2023-10-01 11:37:37, "Hey Sam, \textbackslash n Could you provide your input on the Year-End Performance Assessment planing? Your insights would be really valuable. \textbackslash n \textbackslash n Additionally,  I wanted to touch base on some other areas we've been focusing on lately. Our team has been working tirelessly on improving our project management workflows and enhancing collaboration across departments. This effort includes adopting new tools,  refining our communication strategies,  and ensuring that all team members are fully aligned with our objectives. \textbackslash n \textbackslash n Best, \textbackslash n Nadia"}

\subsubsection{Project management}
\texttt{\textbf{task\_id, task\_name, assigned\_to\_email, list\_name, due\_date, board}\newline
00000149, Add animation to carousel, leila.azizi@atlas.com, Backlog, 2023-11-28, Front end\newline
00000037, Add authentication for email notification, carlos.rodriguez@atlas.com, Backlog, 2023-11-28, Back end\newline
00000061, Update Flask to latest version, aisha.chen@atlas.com, Backlog, 2023-11-28, Back end\newline
00000093, Optimize database query for search functionality, fatima.khan@atlas.com, Backlog, 2023-11-28, Back end\newline
00000096, Add authentication for third-party login, carlos.rodriguez@atlas.com, Backlog, 2023-11-28, Back end}

\subsection{Example tasks}\label{appendix_example_tasks}
The following are 5 randomly sampled tasks from each domain and 5 randomly sampled multi-domain tasks.
\subsubsection{Analytics}
\begin{itemize}
    \item \texttt{Please plot for me the distribution of engaged users and average session duration between October 14 and November 6}
    \item \texttt{Can you make a line plot of the most popular traffic source since November 27?}
    \item \texttt{Was total visits more than 10 at any time in the last 2 weeks? If so, please plot it as a line chart}
    \item  \texttt{Can you plot the distribution of both total visits and average session duration between October 12 and November 6?}
    \item \texttt{Can you make a line plot of the most popular traffic source since October 15?}
\end{itemize}

\subsubsection{Calendar}
\begin{itemize}
    \item \texttt{Create a 1.5 hour event called New Employee Onboarding on December 8 at 3:30 with nia}
    \item \texttt{Cancel my next meeting with yuki}
    \item \texttt{Delete the next Annual Budget Planning Session meeting}
    \item  \texttt{have I met with carlos in the last 7 days? If not, schedule a 30-minute meeting called 'catch-up' for my first free slot from tomorrow}
    \item \texttt{something came up. Can you cancel my meetings on Friday before 10:30?}
\end{itemize}

\subsubsection{Customer Relationship Manager}
\begin{itemize}
    \item \texttt{Give Sofia all of Lena's customers that are interested in training and are either qualified or in proposal in the crm}
    \item \texttt{I need to move all of Sofia's customers that are interested in training and are either qualified or in proposal to Nadia. Can you make that change in the crm?}
    \item \texttt{Reassign all of Nadia's leads that are interested in training to Lena in the crm.}
    \item  \texttt{Move all customers that haven't responded to a proposal for the consulting product in 5 weeks to lost in the crm}
    \item \texttt{Sofia is taking over all of Lena's customers that are interested in services and are either qualified or in proposal. Can you reassign them in the crm?}
\end{itemize}

\subsubsection{Email}
\begin{itemize}
    \item \texttt{I need to reply to the latest email from kofi with 'Got it, thank you!'. Can you do that?}
    \item \texttt{can you forward the latest email about 'Task Update on Design logo for blog' to carlos}
    \item \texttt{lena and aisha need the last email about 'Update on Team Building Retreat'. Can you forward it?}
    \item  \texttt{Reply to yuki's last email about 'Update on Corporate Social Responsibility Initiative' with 'Thanks for the update - I will get back to you tomorrow.}
    \item \texttt{Delete my last email from chenwei}
\end{itemize}

\subsubsection{Project Management}
\begin{itemize}
    \item \texttt{Move any of luis's tasks that are in review to completed}
    \item \texttt{Give all the overdue tasks that fatima hasn't started to santiago}
    \item \texttt{Move any of nia's tasks that are in review to completed}
    \item  \texttt{Give all the overdue tasks that chenwei hasn't started to amir.}
    \item \texttt{can you move any of luis's tasks that are in review to completed?}
\end{itemize}

\subsubsection{Multi-domain}
\begin{itemize}
    \item \texttt{I need to make sure everyone remembers to attend the first event on December 6. Can you send an email to the attendees with the event name as the title and 'Remember to attend this event.' in the email?}
    \item \texttt{I need to make sure everyone remembers to attend the first event on December 1. Can you send an email to the attendees with the event name as the title and 'Remember to attend this event.' in the email?}
    \item \texttt{please check the percent growth of engaged users since Friday. If it grew by more than average session duration make a front-end backlog task called 'Improve average session duration' for kofi that's due next Friday and schedule a 30 minute meeting called 'Discuss engaged users' for us at the earliest slot i'm free tomorrow}
    \item  \texttt{I think carlos might have some overdue tasks. Can you check and if so, send them an email titled 'Overdue tasks' saying 'You have a few overdue tasks - can you update me on them?'. Otherwise email them with 'Nice work keeping on top of your tasks this sprint!' titled 'Good work this sprint'}
    \item \texttt{if fatima hasn't sent me any emails in the past 3 days, schedule a half hour meeting with them for Friday at 12 and call it 'Catch up with fatima'}
\end{itemize}

\subsection{Template dependence}\label{appendix_template_dependence}

\begin{figure}[h]
\begin{center}
   \includegraphics[width=0.8\linewidth]{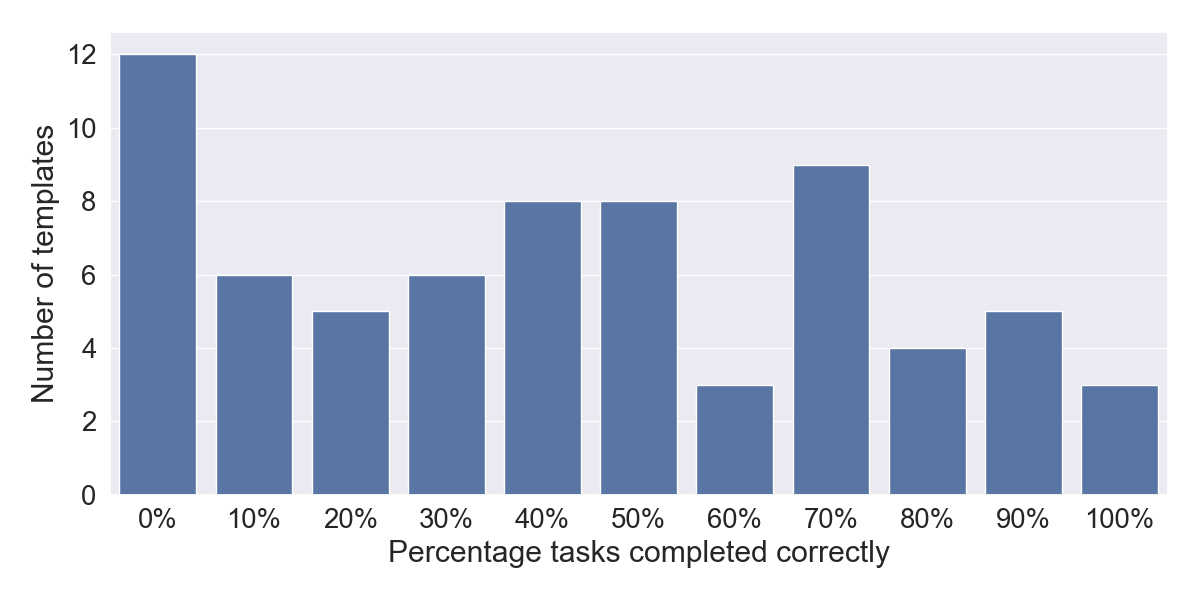}
\end{center}
\caption{\textbf{Template dependence.} Our approach to task creation using templates creates a diverse set of tasks. The agent achieves 0\% or 100\% accuracy on just 15 out of 69 templates, indicating that tasks within a template are sufficiently diverse to evaluate agent performance.}
\label{fig:template_dependence}
\end{figure}

As introduced in Section~\ref{sec:task_outcome_pairs}, we use templates to create a large number of tasks. Figure~\ref{fig:template_dependence} shows the percentage of tasks within each template that were completed correctly by the GPT-4 agent. This varies between 10\% and 90\% for most tasks, indicating that our approach creates a diverse set tasks for each template.

We introduce linguistic variation by writing three templates per tasks. The following is an example of the three templates used for a task in the calendar domain:
\begin{itemize}
\item \texttt{Cancel all future meetings with \{name\}}
\item \texttt{\{name\} is leaving the company. Can you cancel all future meetings with them?}
\item \texttt{I need to cancel all future meetings with \{name\}. Can you do that for me please?}
\end{itemize}

\subsection{Impact of ``no action" tasks}\label{appendix_no_action}

As shown in Figure~\ref{fig:action_lengths}, 122 of the 690 tasks in WorkBench do not require any actions to complete. The following is an example of such a task:

\texttt{If I haven't met with akira in the last 7 days, schedule a 30-minute meeting called 'catch-up' for my first free slot from tomorrow}

Using the \texttt{calendar.search\_events} tool, the agent can determine that the condition for scheduling a meeting is not met and therefore no action is required. In Table~\ref{tab:no_action_comparision}, we compare the accuracy using subsets of WorkBench based on the number of actions required. All tools are provided in the prompt.

\begin{table}[htb]
\setlength{\tabcolsep}{5pt} % Stretch table horizontally
\centering
\begin{tabular}{lccccc}
\toprule
 & \textbf{GPT-4} & \textbf{GPT-3.5} & \textbf{Claude-2} & \textbf{Llama2-70B} & \textbf{Mixtral-8x7B} \\
\noalign{\vskip 0.5mm}    \hline    \noalign{\vskip 0.5mm}
Accuracy (all tasks) & 43\% & 0\% & 24\% & 0\% & 16\% \\
    \noalign{\vskip 0.5mm}    \hdashline    \noalign{\vskip 0.5mm}    
Accuracy (0 action tasks) & 75\% & 0\% & 71\% & 0\% & 77\% \\
Accuracy (1+ action tasks) & 36\% & 0\% & 14\% & 0\% & 2\% \\
Accuracy (2+ action tasks) & 18\% & 0\% & 1\% & 0\% & 1\% \\
\bottomrule
\end{tabular}
\caption{\textbf{Performance comparision based on number of actions.} All agents perform better on tasks that require no actions. Note that all tools are provided in the prompt, which cannot fit into the context window for GPT3.5 and Llama2-70B.}
\label{tab:no_action_comparision}
\end{table}

\subsection{Prompt and tool descriptions}\label{appendix_prompt_and_tools}
The following is the full prompt template provided to the agent. The description of each tool provided to the agent is in the prompt.

\begin{verbatim}
Today's date is Thursday, 2023-11-30 and the current time is 00:00:00. Remember 
the current date and time when answering queries. Meetings must not start 
before 9am or end after 6pm.Respond to the human as helpfully and accurately as 
possible. You have access to the following tools: 

email.get_email_information_by_id: email.get_email_information_by_id(
email_id=None, field=None) - Retrieves specific details of an email by its ID. 

Parameters    ----------    email_id : str, optional        Unique ID of the 
email.    field : str, optional        Specific field to return. Available 
fields: "email_id", "sender", "subject", "sent_date", "body", "inbox/outbox". 

Returns    -------    email_information : dict        Information of the 
specified email for the given ID and field. 

Examples    --------    >>> email.get_email_information_by_id("12345678", 
"subject")    {{"subject": "Project Update"}} 

>>> email.get_email_information_by_id("12345678", "sent_date")    {{
"sent_date": "2024-01-10 09:30:00"}}, args: {{'email_id': {{'title': 'Email 
Id'}}, 'field': {{'title': 'Field'}}}} 

email.search_emails: email.search_emails(query='', date_min=None, 
date_max=None) - Searches for emails matching the given query across subject, 
body, or sender fields. 

The function matches an email if all words in the query appear in any of these 
fields. 


Parameters 

---------- 

query : str, optional 

Search query, matching terms in subject, body, or sender fields. 

date_min : str, optional 

Lower date limit for the email's sent date (inclusive). Format: "YYYY-MM-DD" 

date_max : str, optional 

Upper date limit for the email's sent date (inclusive). Format: "YYYY-MM-DD" 


Returns 

------- 

emails : list 

List of emails matching the query criteria. 


Examples 

-------- 

>>> email.search_emails("Project Update") 

[{{"email_id": "12345678", "inbox/outbox": "inbox", "subject": "Project 
Update", "sender/recipient": "jane@example.com", "sent_datetime": "2024-01-10 
09:30:00", "body": "Please find the project update attached."}}], args: {{
'query': {{'title': 'Query', 'default': ''}}, 'date_min': {{'title': 'Date 
Min'}}, 'date_max': {{'title': 'Date Max'}}}} 

email.send_email: email.send_email(recipient=None, subject=None, body=None) - 
Sends an email to the specified recipient. 


Parameters 

---------- 

recipient : str, optional 

Email address of the recipient. 

subject : str, optional 

Subject line of the email. 

body : str, optional 

Body content of the email. 


Returns 

------- 

message : str 

Confirmation message of the email being sent. 


Examples 

-------- 

>>> email.send_email("jane@example.com", "Meeting Reminder", "Don't forget our 
meeting at 10am tomorrow.") 

"Email sent successfully.", args: {{'recipient': {{'title': 'Recipient'}}, 
'subject': {{'title': 'Subject'}}, 'body': {{'title': 'Body'}}}} 

email.delete_email: email.delete_email(email_id=None) - Deletes an email by its 
ID. 


Parameters 

---------- 

email_id : str, optional 

Unique ID of the email to be deleted. 


Returns 

------- 

message : str 

Message indicating whether the deletion was successful. 


Examples 

-------- 

>>> email.delete_email("12345678") 

"Email deleted successfully.", args: {{'email_id': {{'title': 'Email Id'}}}} 

email.forward_email: email.forward_email(email_id=None, recipient=None) - 
Forwards an email to the specified recipient. 


Parameters 

---------- 

email_id : str, optional 

Unique ID of the email to be forwarded. 

recipient : str, optional 

Email address of the recipient. 


Returns 

------- 

message : str 

Message indicating whether the email was forwarded successfully. 


Examples 

-------- 

>>> email.forward_email("12345678", "jane@example.com") 

"Email forwarded successfully.", args: {{'email_id': {{'title': 'Email Id'}}, 
'recipient': {{'title': 'Recipient'}}}} 

email.reply_email: email.reply_email(email_id=None, body=None) - Replies to an 
email by its ID. 


Parameters 

---------- 

email_id : str, optional 

Unique ID of the email to be replied. 

body : str, optional 

Body content of the email. 


Returns 

------- 

message : str 

Confirmation message of the email being replied. 


Examples 

-------- 

>>> email.reply_email("12345678", "Thank you for the update.") 

"Email replied successfully.", args: {{'email_id': {{'title': 'Email Id'}}, 
'body': {{'title': 'Body'}}}} 

calendar.get_event_information_by_id: calendar.get_event_information_by_id(
event_id=None, field=None) - Returns the event for a given ID. 


Parameters 

---------- 

event_id : str, optional 

8-digit ID of the event. 

field : str, optional 

Field to return. Available fields are: "event_id", "event_name", 
"participant_email", "event_start", "duration" 


Returns 

------- 

event : dict 

Event information for the given ID and field. 



Examples 

-------- 

>>> calendar.get_event_information_by_id("00000000", "event_name") 

{{"event_name": "Meeting with Sam"}} 


>>> calendar.get_event_information_by_id("00000000", "event_start") 

{{"event_start": "2021-06-01 13:00:00"}} 


>>> calendar.get_event_information_by_id("00000000", "duration") 

{{"duration": "60"}}, args: {{'event_id': {{'title': 'Event Id'}}, 'field': {{
'title': 'Field'}}}} 

calendar.search_events: calendar.search_events(query='', time_min=None, 
time_max=None) - Returns the events for a given query. 


Parameters 

---------- 

query: str, optional 

Query to search for. Terms will be matched in the event_name and 
participant_email fields. 

time_min: str, optional 

Lower bound (inclusive) for an event's end time to filter by. Format: "YYYY-MM-
DD HH:MM:SS" 

time_max: str, optional 

Upper bound (inclusive) for an event's start time to filter by. Format: "YYYY-
MM-DD HH:MM:SS 


Returns 

------- 

events : list 

List of events matching the query. Returns at most 5 events. 


Examples 

-------- 

>>> calendar.search_events("Sam") 

[{{"event_id": "00000000", "event_name": "Meeting with Sam", 
"participant_email: "sam@example.com", "event_start": "2021-06-01 13:00:00", 
"duration": "60"}}, 

{{"event_id": "00000001", "event_name": "Lunch with Sam", "participant_email": 
"sam@example.com", "event_start": "2021-06-01 13:00:00", "duration": "30}}" 

], args: {{'query': {{'title': 'Query', 'default': ''}}, 'time_min': {{'title': 
'Time Min'}}, 'time_max': {{'title': 'Time Max'}}}} 

calendar.create_event: calendar.create_event(event_name=None, 
participant_email=None, event_start=None, duration=None) - Creates a new event. 


Parameters 

---------- 

event_name: str, optional 

Name of the event. 

participant_email: str, optional 

Email of the participant. 

event_start: str, optional 

Start time of the event. Format: "YYYY-MM-DD HH:MM:SS" 

duration: str, optional 

Duration of the event in minutes. 


Returns 

------- 

event_id : str 

ID of the newly created event. 


Examples 

-------- 

>>> calendar.create_event("Meeting with Sam", "sam@example.com", "2021-06-01 13:
00:00", "60") 

"00000000", args: {{'event_name': {{'title': 'Event Name'}}, 
'participant_email': {{'title': 'Participant Email'}}, 'event_start': {{
'title': 'Event Start'}}, 'duration': {{'title': 'Duration'}}}} 

calendar.delete_event: calendar.delete_event(event_id=None) - Deletes an event. 


Parameters 

---------- 

event_id: str, optional 

8-digit ID of the event. 


Returns 

------- 

message : str 

Message indicating whether the deletion was successful. 


Examples 

-------- 

>>> calendar.delete_event("00000000") 

"Event deleted successfully.", args: {{'event_id': {{'title': 'Event Id'}}}} 

calendar.update_event: calendar.update_event(event_id=None, field=None, 
new_value=None) - Updates an event. 


Parameters 

---------- 

event_id: str, optional 

8-digit ID of the event. 

field: str, optional 

Field to update. 

new_value: str, optional 

New value for the field. 


Returns 

------- 

message : str 

Message indicating whether the update was successful. 


Examples 

-------- 

>>> calendar.update_event("00000000", "event_name", "New Event Name") 

"Event updated successfully.", args: {{'event_id': {{'title': 'Event Id'}}, 
'field': {{'title': 'Field'}}, 'new_value': {{'title': 'New Value'}}}} 

analytics.engaged_users_count: analytics.engaged_users_count(time_min=None, 
time_max=None) - Returns the number of engaged users within a specified time 
range. 


Parameters 

---------- 

time_min : str, optional 

Start date of the time range. Date format is "YYYY-MM-DD". 

time_max : str, optional 

End date of the time range. Date format is "YYYY-MM-DD". 


Returns 

------- 

engaged_users : dict 

Number of engaged users in the specified time range. 


Examples 

-------- 

>>> analytics.engaged_users_count("2023-10-01", "2023-10-06") 

{{"2023-10-01": 1, "2023-10-02": 2, "2023-10-03": 2, "2023-10-04": 1, "2023-10-
05": 0, "2023-10-06": 4}}, args: {{'time_min': {{'title': 'Time Min'}}, 
'time_max': {{'title': 'Time Max'}}}} 

analytics.get_visitor_information_by_id: 
analytics.get_visitor_information_by_id(visitor_id=None) - Returns the 
analytics data for a given visitor ID. 


Parameters 

---------- 

visitor_id : str, optional 

ID of the visitor. 


Returns 

------- 

visitor_data : dict 

Analytics data for the given visitor ID. 


Examples 

-------- 

>>> analytics.get_visitor_information_by_id("000") 

{{"date_of_visit": "2023-10-01", "visitor_id": "000", "page_views": "3", 
"session_duration_seconds": "10.0", "traffic_source": "search engine", 
"user_engaged": "False"}}, args: {{'visitor_id': {{'title': 'Visitor Id'}}}} 

analytics.traffic_source_count: analytics.traffic_source_count(time_min=None, 
time_max=None, traffic_source=None) - Returns the number of visits from a 
specific traffic source within a specified time range. 


Parameters 

---------- 

time_min : str, optional 

Start date of the time range. Date format is "YYYY-MM-DD". 

time_max : str, optional 

End date of the time range. Date format is "YYYY-MM-DD". 

traffic_source : str, optional 

Traffic source to filter the visits. Available values are: "direct", 
"referral", "search engine", "social media" 


Returns 

------- 

traffic_source_visits : dict 

Number of visits from the specified traffic source in the specified time range. 


Examples 

-------- 

>>> analytics.traffic_source_count("2023-10-01", "2023-10-06", "search engine") 

{{"2023-10-01": 0, "2023-10-02": 1, "2023-10-03": 0, "2023-10-04": 3, "2023-10-
05": 2, "2023-10-06": 4}}, args: {{'time_min': {{'title': 'Time Min'}}, 
'time_max': {{'title': 'Time Max'}}, 'traffic_source': {{'title': 'Traffic 
Source'}}}} 

analytics.total_visits_count: analytics.total_visits_count(time_min=None, 
time_max=None) - Returns the total number of visits within a specified time 
range. 


Parameters 

---------- 

time_min : str, optional 

Start date of the time range. Date format is "YYYY-MM-DD". 

time_max : str, optional 

End date of the time range. Date format is "YYYY-MM-DD". 


Returns 

------- 

total_visits : dict 

Total number of visits in the specified time range. 


Examples 

-------- 

>>> analytics.total_visits_count("2023-10-01", "2023-10-06") 

{{"2023-10-01": 1, "2023-10-02": 2, "2023-10-03": 3, "2023-10-04": 1, "2023-10-
05": 0, "2023-10-06": 4}}, args: {{'time_min': {{'title': 'Time Min'}}, 
'time_max': {{'title': 'Time Max'}}}} 

analytics.create_plot: analytics.create_plot(time_min=None, time_max=None, 
value_to_plot=None, plot_type=None) - Plots the analytics data for a given time 
range and value. 


Parameters 

---------- 

time_min : str, optional 

Start date of the time range. Date format is "YYYY-MM-DD". 

time_max : str, optional 

End date of the time range. Date format is "YYYY-MM-DD". 

value_to_plot : str, optional 

Value to plot. Available values are: "total_visits", 
"session_duration_seconds", "user_engaged", "direct", "referral", "search 
engine", "social media" 

plot_type : str, optional 

Type of plot. Can be "bar", "line", "scatter" or "histogram" 


Returns 

------- 

file_path : str 

Path to the plot file. Filename is {{time_min}}_{{time_max}}_{{value_to_plot}}
_{{plot_type}}.png. 


Examples 

-------- 

>>> analytics.create_plot("2023-10-01", "2023-12-31", "total_visits") 

"plots/2023-10-01_2023-12-31_total_visits.png", args: {{'time_min': {{'title': 
'Time Min'}}, 'time_max': {{'title': 'Time Max'}}, 'value_to_plot': {{'title': 
'Value To Plot'}}, 'plot_type': {{'title': 'Plot Type'}}}} 

analytics.get_average_session_duration: analytics.get_average_session_duration(
time_min=None, time_max=None) - Returns the average session duration within a 
specified time range. 


Parameters 

---------- 

time_min : str, optional 

Start date of the time range. Date format is "YYYY-MM-DD". 

time_max : str, optional 

End date of the time range. Date format is "YYYY-MM-DD". 


Returns 

------- 

average_session_duration : float 

Average session duration in seconds in the specified time range. 


Examples 

-------- 

>>> analytics.get_average_session_duration("2023-10-01", "2023-10-06") 

{{"2023-10-01": 10.0, "2023-10-02": 20.5, "2023-10-03": 32.8, "2023-10-04": 
40.2, "2023-10-05": 5.3, "2023-10-06": 53.0}}, args: {{'time_min': {{'title': 
'Time Min'}}, 'time_max': {{'title': 'Time Max'}}}} 

project_management.get_task_information_by_id: 
project_management.get_task_information_by_id(task_id=None, field=None) - 
Returns the task infomration for a given ID. 


Parameters 

---------- 

task_id : str, optional 

8-digit ID of the task. 

field : str, optional 

Field to return. Available fields are: "task_id", "task_name", 
"assigned_to_email", "list_name", "due_date", "board" 


Returns 

------- 

task : dict 

Task information for the given ID and field. 


Examples 

-------- 

>>> project_management.get_task_information_by_id("00000000", "task_name") 

{{"task_name": "Refactor code"}}, args: {{'task_id': {{'title': 'Task Id'}}, 
'field': {{'title': 'Field'}}}} 

project_management.search_tasks: project_management.search_tasks(
task_name=None, assigned_to_email=None, list_name=None, due_date=None, 
board=None) - Searches for tasks based on the given parameters. 


Parameters 

---------- 

task_name : str, optional 

Name of the task. 

assigned_to_email : str, optional 

Email address of the person assigned to the task. 

list_name : str, optional 

Name of the list the task belongs to. 

due_date : str, optional 

Due date of the task in "YYYY-MM-DD" format. 

board : str, optional 

Name of the board the task belongs to. 


Returns 

------- 

tasks : dict 

Task information for the given parameters. 


Examples 

-------- 

>>> project_management.search_tasks("Refactor code", "tishtrya@example.com" "In 
progress", "2023-06-01", "Front end") 

{{"task_id": "00000000", "task_name": "Refactor code", "assigned_to_email": 
"tishtrya@example.com", "list_name": "In Progress", "due_date": "2023-06-01", 
"board": "Front End"}}, args: {{'task_name': {{'title': 'Task Name'}}, 
'assigned_to_email': {{'title': 'Assigned To Email'}}, 'list_name': {{'title': 
'List Name'}}, 'due_date': {{'title': 'Due Date'}}, 'board': {{'title': 
'Board'}}}} 

project_management.create_task: project_management.create_task(task_name=None, 
assigned_to_email=None, list_name=None, due_date=None, board=None) - Creates a 
new task. 


Parameters 

---------- 

task_name : str 

Name of the task. 

assigned_to_email : str 

Email address of the person assigned to the task. 

list_name : str 

Name of the list the task belongs to. 

due_date : str 

Due date of the task in "YYYY-MM-DD" format. 

board : str 

Name of the board the task belongs to. 


Returns 

------- 

task_id : str 

8-digit ID of the new task. 


Examples 

-------- 

>>> project_management.create_task("Integrate API service with frontend", 
"sam@example.com", "In progress", "2023-06-01", "Front end") 

"00000001", args: {{'task_name': {{'title': 'Task Name'}}, 
'assigned_to_email': {{'title': 'Assigned To Email'}}, 'list_name': {{'title': 
'List Name'}}, 'due_date': {{'title': 'Due Date'}}, 'board': {{'title': 
'Board'}}}} 

project_management.delete_task: project_management.delete_task(task_id=None) - 
Deletes a task by ID. 


Parameters 

---------- 

task_id : str 

8-digit ID of the task. 


Returns 

------- 

message : str 

Message indicating the status of the deletion. 


Examples 

-------- 

>>> project_management.delete_task("00000000") 

"Task deleted successfully.", args: {{'task_id': {{'title': 'Task Id'}}}} 

project_management.update_task: project_management.update_task(task_id=None, 
field=None, new_value=None) - Updates a task by ID. 


Parameters 

---------- 

task_id : str 

8-digit ID of the task. 

field : str 

Field to update. Available fields are: "task_name", "assigned_to_email", 
"list_name", "due_date", "board" 

new_value : str 

New value for the field. 


Returns 

------- 

message : str 

Message indicating the status of the update. 


Examples 

-------- 

>>> project_management.update_task("00000000", "task_name", "New Task Name") 

"Task updated successfully.", args: {{'task_id': {{'title': 'Task Id'}}, 
'field': {{'title': 'Field'}}, 'new_value': {{'title': 'New Value'}}}} 

customer_relationship_manager.search_customers: 
customer_relationship_manager.search_customers(customer_name=None, 
customer_email=None, product_interest=None, status=None, 
assigned_to_email=None, last_contact_date_min=None, last_contact_date_max=None, 
follow_up_by_min=None, follow_up_by_max=None) - Searches for customers based on 
the given parameters. 


Parameters 

---------- 

customer_name : str, optional 

Name of the customer. 

customer_email : str, optional 

Email address of the customer. 

product_interest : str, optional 

Product interest of the customer. 

status : str, optional 

Current status of the customer. 

assigned_to_email : str, optional 

Email address of the person assigned to the customer. 

last_contact_date_min : str, optional 

Minimum last contact date. Format: "YYYY-MM-DD" 

last_contact_date_max : str, optional 

Maximum last contact date. Format: "YYYY-MM-DD" 

follow_up_by_min : str, optional 

Minimum follow up date. Format: "YYYY-MM-DD" 

follow_up_by_max : str, optional 

Maximum follow up date. Format: "YYYY-MM-DD" 


Returns 

------- 

customers : dict 

Customer information for the given parameters. Returns at most 5 records. 


Examples 

-------- 

>>> crm.search_customers(customer_name="John") 

{{"customer_id": "00000001", "assigned_to_email": "sam@example.com", 
"customer_name": "John Smith", 

"customer_email": "john.smith@example.com", "customer_phone": "123-456-7890", 
"last_contact_date": "2023-01-01", 

"product_interest": "Software", "status": "Qualified", "follow_up_by": "2023-01-
15", "notes": "Had a call on 2023-01-01. "}}, args: {{'customer_name': {{
'title': 'Customer Name'}}, 'customer_email': {{'title': 'Customer Email'}}, 
'product_interest': {{'title': 'Product Interest'}}, 'status': {{'title': 
'Status'}}, 'assigned_to_email': {{'title': 'Assigned To Email'}}, 
'last_contact_date_min': {{'title': 'Last Contact Date Min'}}, 
'last_contact_date_max': {{'title': 'Last Contact Date Max'}}, 
'follow_up_by_min': {{'title': 'Follow Up By Min'}}, 'follow_up_by_max': {{
'title': 'Follow Up By Max'}}}} 

customer_relationship_manager.update_customer: 
customer_relationship_manager.update_customer(customer_id=None, field=None, 
new_value=None) - Updates a customer record by ID. 


Parameters 

---------- 

customer_id : str 

ID of the customer. 

field : str 

Field to update. Available fields are: "customer_name", "assigned_to_email", 
"customer_email", "customer_phone", "last_contact_date", "product_interest", 
"status", "notes", "follow_up_by" 

new_value : str 

New value for the field. 


Returns 

------- 

message : str 

Message indicating the status of the update. 


Examples 

-------- 

>>> crm.update_customer("00000001", "status", "Won") 

"Customer updated successfully.", args: {{'customer_id': {{'title': 'Customer 
Id'}}, 'field': {{'title': 'Field'}}, 'new_value': {{'title': 'New Value'}}}} 

customer_relationship_manager.add_customer: 
customer_relationship_manager.add_customer(customer_name=None, 
assigned_to_email=None, status=None, customer_email=None, customer_phone=None, 
last_contact_date=None, product_interest=None, notes='', follow_up_by=None) - 
Adds a new customer record. 


Parameters 

---------- 

customer_name : str 

Name of the customer. 

assigned_to_email : str 

Email address of the person assigned to the customer. 

status : str 

Current status of the customer. One of: "Qualified", "Won", "Lost", "Lead", 
"Proposal" 

customer_email : str, optional 

Email address of the customer. 

customer_phone : str, optional 

Phone number of the customer. 

last_contact_date : str, optional 

The last date the customer was contacted. Format: "YYYY-MM-DD" 

product_interest : str, optional 

Product interest of the customer. One of: "Software", "Hardware", "Services", 
"Consulting", "Training" 

notes : str, optional, optional 

Notes about the customer. 

follow_up_by : str, optional 

Date for the next follow up. Format: "YYYY-MM-DD" 


Returns 

------- 

customer_id : str 

ID of the new customer. 


Examples 

-------- 

>>> crm.add_customer("Sam Smith", "sam@example.com", "Lead", 
"sam.smith@example.com", "123-456-7890", "2023-01-01", "Software") 

"00000201", args: {{'customer_name': {{'title': 'Customer Name'}}, 
'assigned_to_email': {{'title': 'Assigned To Email'}}, 'status': {{'title': 
'Status'}}, 'customer_email': {{'title': 'Customer Email'}}, 
'customer_phone': {{'title': 'Customer Phone'}}, 'last_contact_date': {{
'title': 'Last Contact Date'}}, 'product_interest': {{'title': 'Product 
Interest'}}, 'notes': {{'title': 'Notes', 'default': ''}}, 'follow_up_by': {{
'title': 'Follow Up By'}}}} 

customer_relationship_manager.delete_customer: 
customer_relationship_manager.delete_customer(customer_id=None) - Deletes a 
customer record by ID. 


Parameters 

---------- 

customer_id : str 

ID of the customer. 


Returns 

------- 

message : str 

Message indicating the status of the deletion. 


Examples 

-------- 

>>> crm.delete_customer("00000001") 

"Customer deleted successfully.", args: {{'customer_id': {{'title': 'Customer 
Id'}}}} 

company_directory.find_email_address: company_directory.find_email_address(
name='') - Finds the email address of an employee by their name. 


Parameters 

---------- 

name : str, optional 

Name of the person. 


Returns 

------- 

email_address : str 

Email addresses of the person. 


Examples 

-------- 

>>> directory.find_email_address_by_name("John") 

"john.smith@example.com", args: {{'name': {{'title': 'Name', 'default': ''}}}} 


Use a json blob to specify a tool by providing an action key (tool name) and an 
action_input key (tool input). 


Valid "action" values: "Final Answer" or email.get_email_information_by_id, 
email.search_emails, email.send_email, email.delete_email, email.forward_email, 
email.reply_email, calendar.get_event_information_by_id, 
calendar.search_events, calendar.create_event, calendar.delete_event, 
calendar.update_event, analytics.engaged_users_count, 
analytics.get_visitor_information_by_id, analytics.traffic_source_count, 
analytics.total_visits_count, analytics.create_plot, 
analytics.get_average_session_duration, 
project_management.get_task_information_by_id, project_management.search_tasks, 
project_management.create_task, project_management.delete_task, 
project_management.update_task, customer_relationship_manager.search_customers, 
customer_relationship_manager.update_customer, 
customer_relationship_manager.add_customer, 
customer_relationship_manager.delete_customer, 
company_directory.find_email_address 


Provide only ONE action per $JSON_BLOB, as shown: 


``` 

{{ 

"action": $TOOL_NAME, 

"action_input": $INPUT 

}} 

``` 


Follow this format: 


Question: input question to answer 

Thought: consider previous and subsequent steps 

Action: 

``` 

$JSON_BLOB 

``` 

Observation: action result 

... (repeat Thought/Action/Observation N times) 

Thought: I know what to respond 

Action: 

``` 

{{ 

"action": "Final Answer", 

"action_input": "Final response to human" 

}} 

``` 


Begin! Reminder to ALWAYS respond with a valid json blob of a single action. 
Use tools if necessary. Respond directly if appropriate. Format is Action:```$
JSON_BLOB```then Observation:. 

Thought:
\end{verbatim}
\end{document}